\documentclass[sigconf,nonacm,screen=true]{aamas}

\usepackage{xcolor}
\definecolor{AAMASLav}{HTML}{0161FF}
\definecolor{mybluebox}{HTML}{edf3fd} 

\makeatletter
\AtBeginDocument{
  \hypersetup{
    colorlinks=true,
    allcolors=AAMASLav
  }
  \def\@linkcolor{AAMASLav}%
  \def\@anchorcolor{AAMASLav}%
  \def\@citecolor{AAMASLav}%
  \def\@filecolor{AAMASLav}%
  \def\@urlcolor{AAMASLav}%
  \def\@menucolor{AAMASLav}%
  \def\@pagecolor{AAMASLav}%
}
\makeatother

\usepackage{xcolor}
\definecolor{AAMASBrand}{HTML}{0161FF}

\makeatletter
\AtBeginDocument{%
  \let\AAMAS@old@secfont\@secfont
  \def\@secfont{\color{AAMASBrand}\AAMAS@old@secfont}

  \let\AAMAS@old@subsecfont\@subsecfont
  \def\@subsecfont{\color{AAMASBrand}\AAMAS@old@subsecfont}

  \let\AAMAS@old@subsubsecfont\@subsubsecfont
  \def\@subsubsecfont{\color{AAMASBrand}\AAMAS@old@subsubsecfont}

  \let\AAMAS@old@parfont\@parfont
  \def\@parfont{\color{AAMASBrand}\AAMAS@old@parfont}
}
\makeatother

\usepackage{balance}
\usepackage{xcolor}
\usepackage{color}
\usepackage{tcolorbox}
\usepackage{pifont}%
\newcommand{\cmark}{\ding{51}}%
\newcommand{\xmark}{\ding{55}}%
\usepackage{etoolbox}
\usepackage{balance}
\usepackage{algorithm}
\usepackage{algorithmic}
\usepackage{booktabs}
\usepackage{makecell}
\usepackage{tcolorbox}
\usepackage{xcolor}
\usepackage{amsmath}
\usepackage{array}
\usepackage{amsfonts} 
\usepackage{hyperref}
\usepackage{makecell}
\usepackage{adjustbox}
\setcopyright{none}

\makeatletter
\AtBeginDocument{%
  \let\AAMAS@old@titlefont\@titlefont
  \def\@titlefont{\color{AAMASBrand}\AAMAS@old@titlefont}
}
\makeatother

\makeatletter
\AtBeginDocument{%
  \@ifundefined{@authorfont}{}{%
    \let\A@old@authorfont\@authorfont
    \def\@authorfont{\color{black}\A@old@authorfont}%
  }%
  \@ifundefined{@affiliationfont}{}{%
    \let\A@old@affiliationfont\@affiliationfont
    \def\@affiliationfont{\color{black}\A@old@affiliationfont}%
  }%
  \@ifundefined{@emailfont}{}{%
    \let\A@old@emailfont\@emailfont
    \def\@emailfont{\color{black}\A@old@emailfont}%
  }%
  \@ifundefined{@authornotefont}{}{%
    \let\A@old@authornotefont\@authornotefont
    \def\@authornotefont{\color{black}\A@old@authornotefont}%
  }%
}
\makeatother

\title[Don’t Blind Your VLA]{%
  \texorpdfstring{\textcolor{AAMASBrand}{Don’t Blind Your VLA:\\
  Aligning Visual Representations for OOD Generalization}}%
  {Don’t Blind Your VLA: Aligning Visual Representations for OOD Generalization}%
}

\author{Nikita Kachaev}
\affiliation{
  \institution{Cognitive AI Lab}
  \city{Moscow}
  \country{Russia}
}

\author{Mikhail Kolosov}
\affiliation{
  \institution{IAI MIPT}
  \city{Moscow}
  \country{Russia}
}

\author{Daniil Zelezetsky}
\affiliation{
  \institution{IAI MIPT}
  \city{Moscow}
  \country{Russia}}
  
\author{Alexey K. Kovalev}
\affiliation{
  \institution{Cognitive AI Lab, IAI MIPT}
  \city{Moscow}
  \country{Russia}}

\author{Aleksandr I. Panov}
\affiliation{
  \institution{Cognitive AI Lab, IAI MIPT}
  \city{Moscow}
  \country{Russia}}

\begin{abstract}

The growing success of Vision-Language-Action (VLA) models stems from the promise that pretrained Vision-Language Models (VLMs) can endow agents with transferable world knowledge and vision-language (VL) grounding, laying a foundation for action models with broader generalization. Yet when these VLMs are adapted to the action modality, it remains unclear to what extent their original VL representations and knowledge are preserved. In this work, we conduct a systematic study of representation retention during VLA fine-tuning, showing that naive action fine-tuning leads to degradation of visual representations. To characterize and measure these effects, we probe VLA's hidden representations and analyze attention maps, further, we design a set of targeted tasks and methods that contrast VLA models with their counterpart VLMs, isolating changes in VL capabilities induced by action fine-tuning. We further evaluate a range of strategies for aligning visual representations and introduce a simple yet effective method that mitigates degradation and yields improved generalization to out-of-distribution (OOD) scenarios. Taken together, our analysis clarifies the trade-off between action fine-tuning and the degradation of VL representations and highlights practical approaches to recover inherited VL capabilities. Code is publicly available: %
\href{https://blind-vla-paper.github.io}{\textcolor{AAMASBrand}{\textbf{blind-vla-paper.github.io}}}
\end{abstract}
         
\newcommand{\BibTeX}{\rm B\kern-.05em{\sc i\kern-.025em b}\kern-.08em\TeX}

\usepackage{fancyhdr,xcolor}
\makeatletter
\AtBeginDocument{%
  \fancypagestyle{firstpagestyle}{%
    \fancyhf{}%
    \fancyfoot[L]{\scriptsize\color{gray!60}{Under review}}%
  }%
  \fancypagestyle{standardpagestyle}{%
    \fancyhf{}%
  }%
  \pagestyle{standardpagestyle}%
}
\makeatother

\usepackage[font=small,skip=4pt]{caption}

\begin{document}
\pagestyle{plain} 
\fancyhead{}
\maketitle

\section{Introduction}

Vision–Language Models (VLMs) have demonstrated remarkable success due to their ability to integrate large-scale multimodal datasets, thereby acquiring semantic grounding and generalizable visual-language (VL) representations
 \citep{cosmos_reason, Qwen2.5-VL, InternVL3.5, paligemma, palme, openflamingo2023}. When exposed to novel visual or linguistic contexts, such models exhibit robust cross-modal understanding and compositional perception -- properties that underpin their strong zero and few-shot generalization beyond the training distribution. These advancements have naturally inspired the extension of VLMs toward embodied domains.

\begin{figure}[t]
    \centering
    \includegraphics[width=1\linewidth]{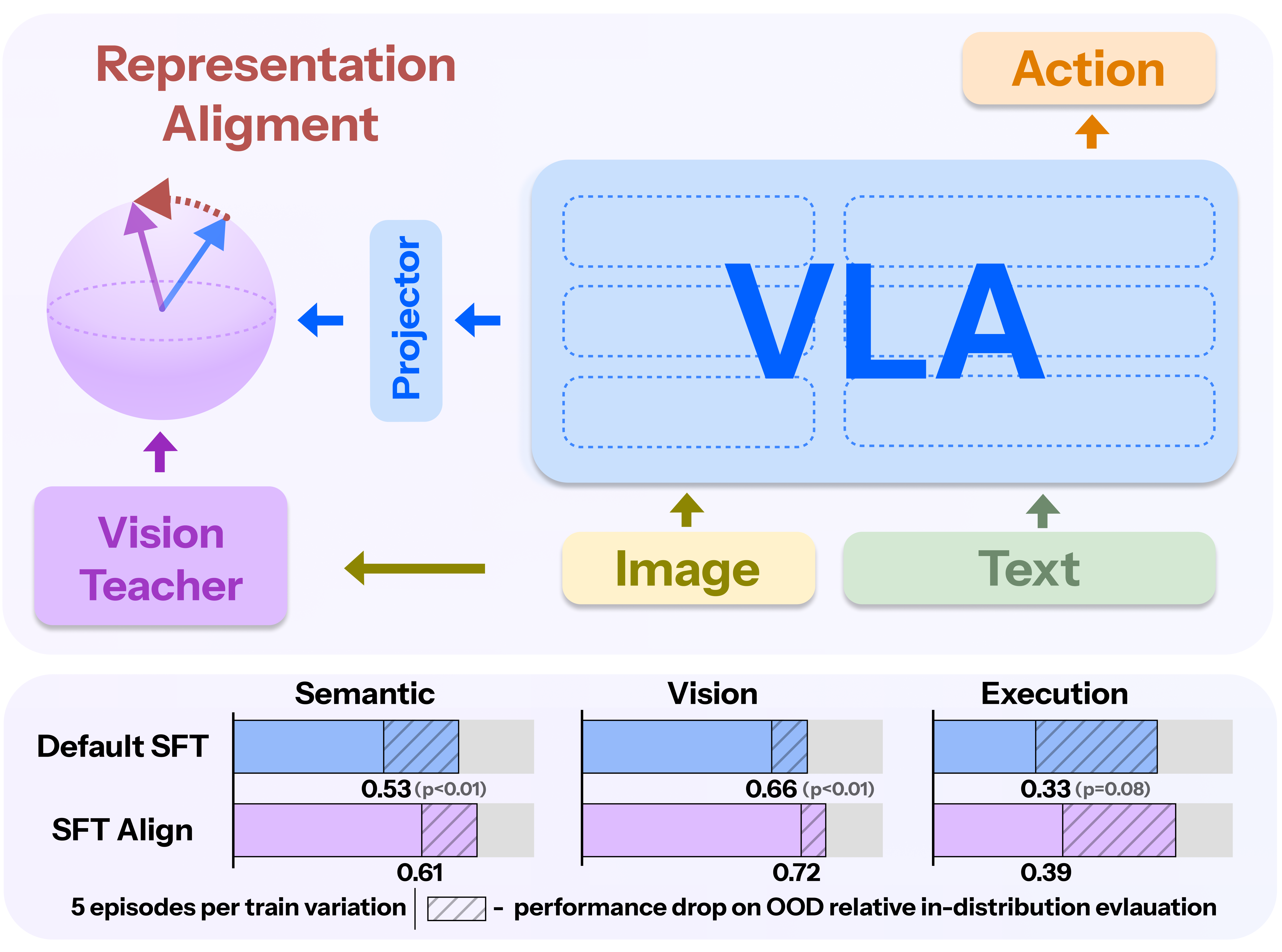}
    \caption{Visual alignment method overview. Mid-level VLA features are projected onto a normalized sphere and aligned with teacher embeddings, preserving visual semantics and improving OOD generalization. Bottom plots show comparison with standard SFT across three generalization axes on the Simpler-based benchmark \citep{liu2025rl4vla}.}
    \label{fig:method_arch}
\end{figure}

Vision–Language–Action (VLA) models represent a prominent direction in this research trajectory. They adapt pretrained VLMs to action prediction tasks in robotic settings, with the goal of leveraging the semantic priors and cognition abilities inherited from large-scale vision–language pretraining. The underlying hypothesis is that, if appropriately adapted, VLA models can transfer the visual–semantic representations of their initial VLM to the action domain, enabling generalization to previously unseen scenes, instructions, and scenarios. However, in practice, adapting VLMs to the action modality often introduces new challenges. Several recent studies \citep{pugacheva2025bringapplesofaimpact, ki, libero, calvin, genaug} have shown that current VLA models struggle to maintain generalization in visually and linguistically complex tasks, raising questions about whether strong VL capabilities of VLMs truly transfer to embodied settings. This issue becomes the most evident during task-specific fine-tuning, where limited data diversity and datasets frequently lead to overfitting \citep{Zang2024Overcoming, staroverov2023fine, Ding2024LOBG, pugacheva2025bringapplesofaimpact, cherepanov2025elmurexternallayermemory}. 

\begin{figure*}
    \centering
    \includegraphics[width=1\linewidth]{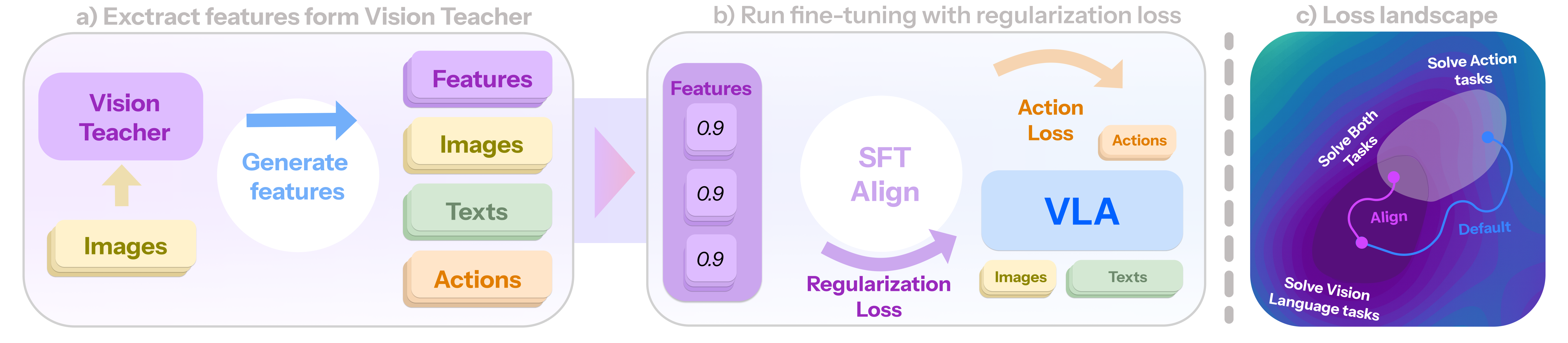}
    \caption{Overview of the proposed method.
    (a, b) Training pipeline with visual alignment loss -- no extra overhead, only precomputed teacher features and a lightweight regularization term during SFT. (c) Conceptual illustration of the loss landscape for VL tasks: the core idea is to optimize the model with respect to the action objective while preserving performance on VL understanding.}
    \label{fig:method_overview}
\end{figure*}

During large-scale robotic pretraining, recent works have attempted to mitigate this degradation by preserving multimodal understanding capabilities. Prior strategies include incorporating auxiliary reasoning objectives \citep{ecot_lite}, applying multimodal co-training on web-scale data \citep{magma}, or freezing pretrained visual–language backbones to preserve VL representations and improve instruction following \citep{groot,ki}. While these approaches help retain vision–language knowledge and improve generalization, they often depend on heavy supervision, high computational cost, or constrained model architecture. Yet, despite these advances at the pretraining stage, there remain no effective methods to address representation degradation during task-specific supervised fine-tuning (SFT) -- the critical phase where VLA models must adapt to certain robotic domains without losing their semantic grounding and VL abilities.

In this work, we adopt a realistic VLA deployment setting: starting from a pretrained VLA and adapting it with limited data for supervised fine-tuning in a chosen embodiment and domain. Under these constraints, we conduct a systematic investigation into the degradation of VL representations and multimodal understanding abilities in VLA models and ask a central question: \textbf{Can we design a simple yet effective method to recover the inherited VL representations during fine-tuning on robotic actions?}

To answer this question, we first examined the attention maps and feature activations of the VLA model in comparison to VLM's across matched image-instruction pairs from the robotics domain.
Our analysis of attention maps revealed that: while the pretrained VLM accurately focuses on task-relevant objects, the fine-tuned VLA models often produce diffuse or misplaced activations, failing to attend to key entities under out-of-distribution (OOD) conditions (\autoref{fig:do_you_see_can}).
Next, we conducted a t-SNE \citep{vandermaaten08a} analysis of intermediate representations across VLM's and VLA's layers, which exposed a clear representation collapse ~\citep{collapse1, collapse2} in VLA models -- indicating that standard action fine-tuning compresses diverse internal features into a narrow representation space, reducing representational diversity and generalization capacity. Next, we propose VL-Think task suite (\autoref{sec:vl_think}) to assess transfer of VL knowledge from VLMs to VLA models, benchmark several strong VLMs and compare OpenVLA–7B \citep{openvla} to its pretrained base (PrismaticVLM \citep{prismaticds}). We observe systematic, domain-specific forgetting after action fine-tuning, indicating that VLAs lose VL knowledge about domains absent from the robotics fine-tuning data.

To address this representational degradation, we introduce a lightweight \textbf{Visual Representation Alignment} method inspired by the \emph{Platonic Representation Hypothesis}~\citep{platonicrepresentationhypothesis}. 
This hypothesis suggests that large vision and language models tend to converge toward a shared latent representation space that encodes general visual and semantic representations across generalist models. Our method explicitly constrains the visual representations of a VLA to remain aligned with a generalist vision model throughout fine-tuning. By maintaining this link, the VLA preserves semantic consistency while adapting its action policy to new tasks. The method adds negligible computational overhead and integrates seamlessly with SFT (\autoref{fig:method_overview}). Extensive experiments on different variations of Simpler~\citep{simpler} benchmark demonstrates that this alignment consistently improves out-of-distribution generalization -- yielding up to a 10\% relative gain over naive SFT (\autoref{tab:ood_performance}).

\textbf{Our key contributions are as follows:}

\begin{enumerate}
    \item We systematically demonstrate that naive VLA fine-tuning induces representation collapse and attention sink relative to their initial VLM. 
    \item We introduce VL-Think, a diagnostic task suite for assessing transfer of VL knowledge from VLMs across VLA models and show that VLA action fine-tuning lead to domain-specific forgetting.
    \item We propose a simple and efficient visual alignment method that anchors the VLA’s vision representations to strong visual teacher features, preserving multimodal understanding and improving OOD generalization without added complexity (\autoref{fig:method_overview}).
\end{enumerate}

Taken together, our findings provide new insights into the trade-off between action fine-tuning and representation degradation in VLA models. They underscore the importance of maintaining visual-language alignment during fine-tuning and provide a practical recipe for building VLAs that do not ``blind'' the pretrained perceptual knowledge they rely upon.

\begin{figure*}[t]
    \centering
    \includegraphics[width=0.9\linewidth]{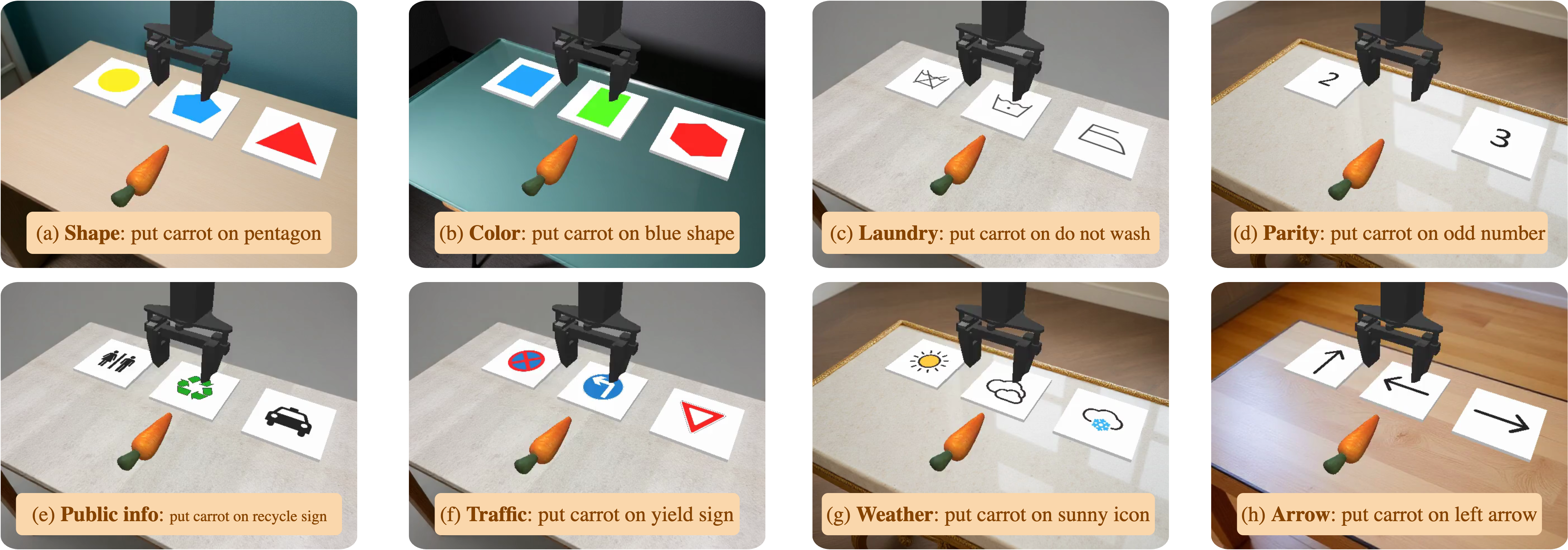}
    \caption{\textbf{VL-Think Task Suite examples.} 
    Each panel illustrates a pick-and-place episode where the agent must place an object on the board matching the instructed concept (e.g., color, number, symbol, or category).}
    \label{fig:vlthink-bench}
\end{figure*}

\section{Related Works}

\subsection{Vision-Language-Action models}

VLA models aim to unify perception, reasoning, and control through large-scale multimodal learning. Early approaches such as RT-1 \citep{rt1} and RT-2 \citep{rt2} demonstrated that scaling VL pretraining to robot data enables generalization across diverse manipulation tasks. Subsequent works -- including OpenVLA \citep{openvla}, Octo \citep{octo}, MolmoAct \citep{moactact}, OneTwoVLA \citep{onetwovla}, and $\pi_0$ \citep{pi0} -- explored large scale robotic pretraining, compact diffusion-based policies, modular reasoning architectures, token-based decision sequencing, and continuous flow-matching policies. Across these models, the shared goal is to couple semantic grounding with low-level motor control in a unified policy, while maintaining efficiency and generalization in real-world settings. A central challenge remains the preservation and retention of VL understanding capabilities during robot fine-tuning.

\subsection{Representation alignment}

Recent studies reveal a consistent pattern: as models scale in parameters, data, and tasks, their representations increasingly align across architectures and modalities. The Platonic Representation Hypothesis \citep{platonicrepresentationhypothesis} frames this as convergence to a shared statistical model of reality, independently trained vision and language encoders show semantically compatible spaces, and large language-free visual models reach CLIP-level performance while naturally aligning with text \citep{visionlanguageencodersrepresent,fan2025scalinglanguagefreevisualrepresentation, fan2025scalinglanguagefreevisualrepresentation}.

Recent representation learning methods reinforce this trend: REPA \citep{repa} aligns diffusion hidden states to strong image encoders (faster training, better ImageNet quality), OLA-VLM \citep{jain2025elevatingvisualperceptionmultimodal} distills multi-teacher targets into intermediate LLM layers via predictive embedding losses, 3DRS \citep{huang2025mllmsneed3dawarerepresentation} injects 3D-aware supervision with multi-view correspondence, and Geometry Forcing \citep{wu2025geometryforcingmarryingvideo} aligns video-diffusion features with a 3D backbone via angular/scale objectives for temporally consistent generations.

\section{Preliminaries}

\noindent\textbf{\textcolor{AAMASBrand}{VLA architecture.}}
Let the input multimodal token sequence to the VLM backbone be
\begin{equation}
x_{1:n} = [x_{1:k},\, x_{k+1:n}].
\end{equation}
where \(x_{1:k}\) correspond to visual tokens
and \(x_{k+1:n}\) correspond to textual instruction tokens.
These tokens are obtained from two encoders:
\begin{equation}
x_{1:k}=E_{\mathrm{image}}(I) \in \mathbb{R}^{k \times d_e},
\qquad
x_{k+1:n}=E_{\mathrm{text}}(\ell) \in \mathbb{R}^{(n-k)\times d_e}.
\end{equation}
where \(E_{\mathrm{image}}\) and \(E_{\mathrm{text}}\) denote the image and text encoders into the common
embedding space of dimension \(d_e\) of the VLA model,
and \(I\) and \(\ell\) are the input image and textual instruction, respectively. The combined sequence \(x_{1:n}\) is processed by a multimodal Transformer backbone
\(B_\theta:\mathbb{R}^{n\times d_e}\to\mathbb{R}^{n\times d_e}\) with \(L\) stacked layers.
Denote the hidden states after layer \(i\) by \(h^i_{1:n}\in\mathbb{R}^{n\times d_e}\). Each layer updates the hidden states using standard self-attention with \(h^{0}_{1:n} = x_{1:n}\).:
\begin{equation}
h^{i}_{1:n} = \mathrm{Attention}(h^{i-1}_{1:n}) + \mathrm{FFN}(h^{i-1}_{1:n}), \qquad i = 1,\dots,L.
\end{equation}

\noindent\textbf{\textcolor{AAMASBrand}{Autoregressive objective.}}
Let \(y_{1:m}\) denote the target output tokens (from the same vocabulary as text tokens).
At the decoding step \(t\), the model conditions on the concatenation of the input and the previously
generated tokens:
\begin{equation}
\tilde{x}^{(t)}_{1:n+t-1} = [\,x_{1:n},\, y_{1:t-1}\,], \qquad
h^L_{1:n+t-1} = B_\theta\!\big(\tilde{x}^{(t)}_{1:n+t-1}\big).
\end{equation}
The Transformer then defines the autoregressive distribution
\begin{equation}
p_\theta\!\big(y_t \mid x_{1:n}, y_{1:t-1}\big)
= \mathrm{softmax}\!\big(W_o\, h^L_{n+t-1}\big)\big[\,y_t\,\big].
\end{equation}
where \(W_o\) is the output projection to the token vocabulary, the causal mask in \(B_\theta\)
ensures that \(h^L_{n+t-1}\) depends only on \(x_{1:n}\) and \(y_{1:t-1}\). Training uses the standard next-token loss:
\begin{equation}
\mathcal{L}_{\mathrm{VLA}}(\theta)
=
\mathbb{E}_{(x,y)\sim\mathcal{D}}\!\left[
-\sum_{j=1}^{n-1} M_j\,\log p_\theta\!\big(y_{j+1}\mid x_{1:j}\big)
\right].
\end{equation}
with mask $M$ selecting target positions (we consider the usual causal language-modeling setup).

\section{VL-Think Task Suite}
\label{sec:vl_think}
    
Current evaluations of VLA models \citep{libero, calvin, mikasa} primarily emphasize task execution under distribution shifts -- such as changes in objects, scenes, recall-based demands or textures but provide little insight into whether the VL capabilities and knowledge inherited from the pretrained VLM are preserved after action fine-tuning. To address this gap, we introduce the \textbf{VL-Think Task Suite}, a diagnostic suite designed to evaluate the transfer of VL capabilities from VLMs to VLAs independently of their low-level control performance. The suite focuses on testing whether a model continues to understand visual symbols, compositional cues, and categorical distinctions that are commonly evaluated in VLM datasets but underrepresented in robotics domain -- rather than whether it can successfully execute grasp or placement actions. We intentionally minimize control complexity to ensure that any observed performance degradation reflects a loss of VL understanding, rather than action execution.

\subsection{Evaluation protocol}
To quantify the gap in VL capabilities, we perform evaluations across both VLA and VLM models. 

\noindent\textbf{\textcolor{AAMASBrand}{VLA evaluation.}} The agent observes RGB frames and language instructions. The success rate is recorded if a well-known object is placed on the correct target board. Since motion complexity is fixed, this directly measures the model’s capacity to ground language in visual categories rather than its manipulation skills. 

\noindent\textbf{\textcolor{AAMASBrand}{VLM evaluation.}} To assess reasoning in robotics setup without actions, the same scenes are presented as static initial images with the probe: \textit{``Do you see the <board\_name>?''. Answer ‘yes’ or ‘no’. If yes, specify where: ‘left’, ‘center’, or ‘right’”}. A response is counted as successful only if both the predicted board and its target location match the ground truth, yielding a success rate that serves as an action-free measure of semantic grounding.

\subsection{VL-Think description}
To reduce the embodiment and setup-specific adaptation bottlenecks, VL-Think Task Suite is based on the realistic Simpler \citep{simpler} benchmark with WidowX-250S arm pick-and-place task. Each episode spawns a single source well-known object (carrot) positioned to yield 100\% grasp reliability and multiple planar ``boards'' textured with abstract categories (e.g., icons, shapes, numerals). A language instruction specifies a single target concept (shape, color, icon class, direction, or parity). The agent succeeds if it places the carrot on the board that matches the instructed concept. By keeping the objects and action complexity fixed, the evaluation isolates VL skills while bounding execution complexity.

The VL-Think suite consists of eight board-selection tasks that probe different aspects of knowledge (see Figure~\ref{fig:vlthink-bench}). In each task, the agent must place the object on the board that matches the instructed concept: \textbf{Shape} -- the board whose graphic is the named geometric shape; e.g., ``Put the object on the star.''), \textbf{Color} -- the board whose shape has the named color; e.g., ``Put the object on the blue shape'', \textbf{Traffic} -- the board depicting one of 24 common traffic signs; e.g., ``the yield sign'', \textbf{Laundry care} -- the board depicting one of 17 standard laundry symbols, e.g., ``Do not bleach'', \textbf{Weather} -- the board depicting one of 9 common weather icons; e.g., ``sunny'', ``cloudy'', \textbf{Directional arrow} -- the board whose arrow points in the named direction: ``up'', ``down'', ``left'', ``right'', \textbf{Public information} -- the board depicting one of 14 public-information signs; e.g., ``no dogs allowed'', and \textbf{Numeral parity} -- the board whose printed numeral matches the requested parity (``odd'' or ``even''); e.g., ``Put the object on the odd number''.

\section{VL representations analysis}

In this section, we ask: what happens to VL representations and knowledge in VLA models after action fine-tuning? Does knowledge transfer from VLMs actually occur, and is strong semantic grounding retained?

To examine how strongly VL representations degrade in VLA models, we conduct complementary analyses. First, we use t-SNE~\citep{JMLR:v9:vandermaaten08a} visualization to assess whether the model preserves a structured and separable latent space for instruction-related tokens. Second, we analyze attention maps to evaluate how accurately the model focuses on objects referenced in the input instruction. Finally, using the VL-Think suite, we assess the transferability of VLM VL skills to VLA policies. Together, these methods provide intuitive and interpretable diagnostics of VL representation degradation and domain forgetting -- revealing whether the model maintains focused visual grounding, coherent latent organization and erodes domain-specific knowledge after action fine-tuning.

\subsection{Attention sink}
\label{sec:attn}

\begin{figure}[t]
    \centering
    \includegraphics[width=\linewidth]{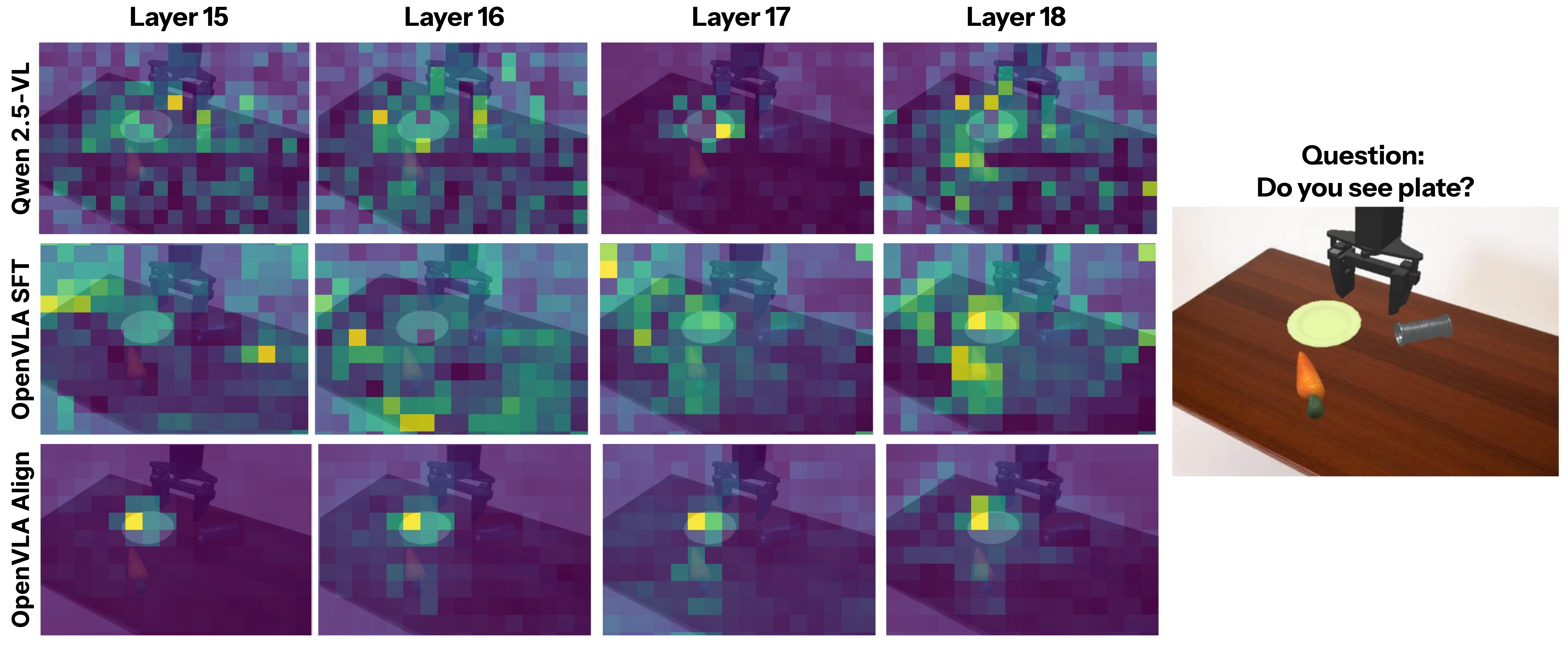}
    \caption{Attention map comparison: the strongest and most semantically grounded attention appears around middle layers. OpenVLA fine-tuned with our proposed method (OpenVLA Align) maintains object-aligned focus in attention maps, while default OpenVLA SFT shows diffused and noisy patterns, indicating loss of visual-language grounding (for more results see Appendix \autoref{fig:do_you_see_canv2}).}
    \label{fig:do_you_see_can}
\end{figure}

To further investigate how fine-tuning affects the VL grounding capabilities of VLA models, we examine their attention maps, which reveal how effectively the model focuses on the object referenced in a textual instruction. This analysis provides a direct probe into how well the model maintains connection between visual and language features. For each model, we visualize the attention maps for visual patch embeddings from the middle layers. Following prior studies \citep{zhang2025mllms}, we observe (\autoref{fig:do_you_see_can}) that the strongest and most semantically meaningful attention patterns typically emerge in the middle transformer layers (layers 14–24), where vision–language fusion is the most active.

Among the evaluated models, Qwen2.5-VL exhibits clear and relevant object-aligned attention, indicating that its attention is precisely localized on the queried object with minimal spatial noise. In contrast, OpenVLA displays substantial degradation in attention quality: the maps become diffuse, noisy, and weakly correlated with the target object indicating attention sink \citep{kang2025toldvisualattentionsink, lin2025boostingmultimodallargelanguage}. Instead of concentrating on relevant image regions, the OpenVLA's attention maps frequently leak into irrelevant background regions or concentrate on distractor objects (for more results see \autoref{app:att_map}). By contrast, our proposed Visual Representation Alignment approach remedies this issue: OpenVLA (Align) trained with it produces crisp, object-centric attention maps (see \autoref{app:att_map} for details).

\label{sec:motivation}
\begin{figure}[t]
    \centering
    \includegraphics[width=0.9\linewidth]{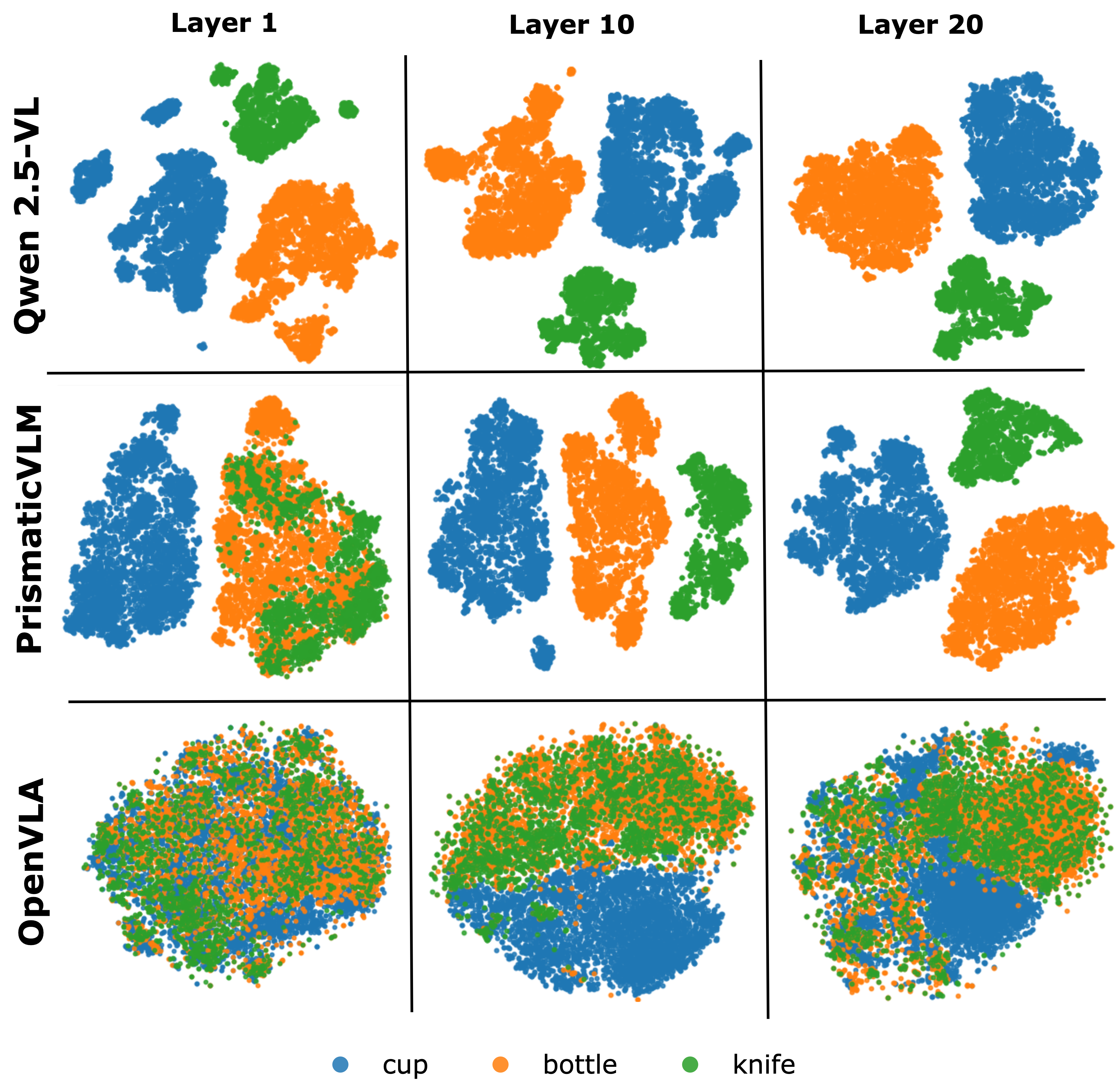}
    \caption{t-SNE visualization of token embeddings for Qwen2.5-VL, PrismaticVLM, and OpenVLA. While PrismaticVLM and Qwen2.5-VL maintains well-separated clusters for target objects, OpenVLA shows huge overlap across classes, indicating that action fine-tuning causes representations collapse.}
    \label{fig:vla-tsne-main}
\end{figure}

\subsection{Representations collapse}
\label{sec:tsne}

To analyze how action fine-tuning affects the internal VL representations of VLA models, we conducted a t-SNE representation probe comparing Qwen2.5-VL \citep{Qwen2.5-VL}, PrismaticVLM \citep{prismaticds}, and OpenVLA \citep{openvla}. This experiment provides a qualitative view of how the semantic structure in the latent space evolves through the action training process. We use the COCO dataset \citep{lin2014microsoft} and select samples from three common household object classes: cup, bottle, and knife. For each image, the model receives a textual query of the form \textit{``Do you see <object\_name>?''}. Then we extract the embedding corresponding to the token \textit{<object\_name>} from transformer layers and then project these embeddings into two dimensions using the t-SNE algorithm. Each point in the visualization is color-coded by its object class, allowing us to observe how distinct or entangled the category clusters become.

\autoref{fig:vla-tsne-main} illustrates this comparison for the middle layers, revealing how the latent space is organized across the different model's layers. In the PrismaticVLM and Qwen2.5-VL, embeddings for the three categories form well-separated clusters reflecting a coherent and semantically organized latent space typical of large-scale VLMs. In contrast, OpenVLA exhibits blurred and overlapping clusters, indicating that fine-tuning for robot control disrupts the structured organization of its inherited representations. This loss of separability corresponds to a phenomenon akin to representation collapse~\citep{collapse1,collapse2}, where previously distinct VL representations converge into less discriminative subspaces. 

\subsection{Domain forgetting in VLA models}
\label{sec:vl_think_eval1}

Using the VL-Think task suite (\autoref{sec:vl_think}), we evaluate VL capabilities across several state-of-the-art VLMs: InternVL3.5 \citep{InternVL3.5}, Ovis2.5 \citep{Ovis2.5}, Qwen2.5-VL \citep{Qwen2.5-VL} and focus on OpenVLA–7B \citep{openvla} versus its pretrained base PrismaticVLM \citep{prismaticds}, which we use as an approximate upper bound. This comparison probes how much VL knowledge and semantic grounding skills persist after action fine-tuning.

Two clear trends emerge. First, strong VLMs achieve high success rate across all domains, reflecting robust semantic grounding. Second, action fine-tuning induces systematic, domain-specific forgetting in VLA models: relative to its pretrained counterpart, OpenVLA–7B exhibits substantial drops in nearly all domains, with the largest declines in symbolic and abstract categories (traffic, arrows, public information, weather). We hypothesize that VLA models lose knowledge about domains that are absent in robotics fine-tuning datasets. The single domain where transfer persists is \textit{Color}: the success rate remains at the level of the initial VLM, likely because color cues are directly useful for control and are implicitly present in robotics datasets.

\section{Method}
\label{sec:method}

Following the \textit{Platonic Representation Hypothesis}~\citep{platonicrepresentationhypothesis},
we assume that high-performing vision, language, and multimodal models
tend to converge toward a shared latent representation space
that captures general semantic and perceptual structure across different modalities.
Each modality provides a distinct but compatible view of this shared space,
encoding complementary aspects of the same underlying VL regularities.
From this perspective, a VLA model can be regarded as a policy that grounds its decision-making in a subset of these multimodal representations.
However, during task-specific fine-tuning,
the policy's internal features may drift away from this generalized representation space,
causing it to lose connection to broad, transferable semantics. To mitigate this effect, we introduce a Visual Representation Alignment objective
that anchors the VLA’s visual representations to a stable external reference
encoding consistent, general-purpose visual semantics (\autoref{fig:method_arch}).

\begin{table*}[t]
\centering
\caption{OOD generalization performance across evaluation environments (mean $\pm$~SD). 
The proposed alignment objective yields consistent gains over SFT and frozen-encoder baselines, indicating enhanced robustness to OOD domain shifts.}
\setlength{\tabcolsep}{3pt}
\renewcommand{\arraystretch}{1.15}
\resizebox{\textwidth}{!}{%
\begin{tabular}{lccccccccccccc}
\toprule
\textbf{Method} &
\multicolumn{5}{c}{\textbf{Semantic}} &
\multicolumn{5}{c}{\textbf{Vision}} &
\multicolumn{3}{c}{\textbf{Execution}} \\
\cmidrule(lr){2-6}
\cmidrule(lr){7-11}
\cmidrule(lr){12-14}
& Carrot & Instruct & MultiCarrot & MultiPlate & Plate
& VisionImg & Tex03 & Tex05 & Whole03 & Whole05
& Position & EEPose & PosChangeTo \\
\midrule
Default &
$0.49{\scriptsize \pm 0.02}$ &
$0.74{\scriptsize \pm 0.02}$ &
$0.28{\scriptsize \pm 0.02}$ &
$0.43{\scriptsize \pm 0.02}$ &
$\underline{0.73{\scriptsize \pm 0.02}}$ &
$0.81{\scriptsize \pm 0.01}$ &
$0.67{\scriptsize \pm 0.01}$ &
$0.55{\scriptsize \pm 0.03}$ &
$0.71{\scriptsize \pm 0.02}$ &
$0.56{\scriptsize \pm 0.01}$ &
$0.43{\scriptsize \pm 0.02}$ &
$0.34{\scriptsize \pm 0.01}$ &
$\boldsymbol{0.23{\scriptsize \pm 0.01}}$ \\
Freeze &
$0.03{\scriptsize \pm 0.01}$ &
$0.05{\scriptsize \pm 0.01}$ &
$0.01{\scriptsize \pm 0.01}$ &
$0.02{\scriptsize \pm 0.01}$ &
$0.03{\scriptsize \pm 0.01}$ &
$0.02{\scriptsize \pm 0.01}$ &
$0.03{\scriptsize \pm 0.01}$ &
$0.01{\scriptsize \pm 0.01}$ &
$0.01{\scriptsize \pm 0.01}$ &
$0.01{\scriptsize \pm 0.01}$ &
$0.03{\scriptsize \pm 0.01}$ &
$0.03{\scriptsize \pm 0.01}$ &
$0.04{\scriptsize \pm 0.01}$ \\
Align (ours) &
$\boldsymbol{0.61{\scriptsize \pm 0.01}}$ &
$\boldsymbol{0.83{\scriptsize \pm 0.03}}$ &
$\boldsymbol{0.35{\scriptsize \pm 0.02}}$ &
$\boldsymbol{0.49{\scriptsize \pm 0.02}}$ &
$\boldsymbol{0.75{\scriptsize \pm 0.01}}$ &
$\boldsymbol{0.86{\scriptsize \pm 0.02}}$ &
$\boldsymbol{0.70{\scriptsize \pm 0.02}}$ &
$\boldsymbol{0.67{\scriptsize \pm 0.02}}$ &
$\boldsymbol{0.79{\scriptsize \pm 0.02}}$ &
$\boldsymbol{0.60{\scriptsize \pm 0.02}}$ &
$\boldsymbol{0.58{\scriptsize \pm 0.02}}$ &
$\boldsymbol{0.38{\scriptsize \pm 0.02}}$ &
$\underline{0.20{\scriptsize \pm 0.03}}$ \\
\bottomrule
\label{tab:ood_performance}
\end{tabular}%
}
\end{table*}

\subsection{Visual representation alignment}
\label{sec:visual_aligment}

We propose a lightweight visual alignment method that recover
generalized and semantically consistent visual representations inside
a VLA model by regularizing its internal embeddings to remain close to those of a frozen, pretrained vision teacher.
In the Platonic interpretation, the teacher encoder provides a more stable
and semantically precise projection of the generalized representation space, while the VLA’s own representations form a task-adapted approximation of this space.
By minimizing their discrepancy, the model is guided back toward a
common semantic structure.

Let $E^{\star}_{\mathrm{img}}$ denote the frozen teacher encoder that produces patch-level features
\begin{equation}
z_{1:k}=E^{\star}_{\mathrm{img}}(I)\in\mathbb{R}^{k\times d_t},
\end{equation}
where each patch embedding $z_{m-1:m}$ captures localized visual semantics within the teacher’s high-level feature space.
Within the VLA model, we select an internal layer $i^\star$ that carries
semantically rich visual information and extract the corresponding vision tokens
$h^{i^\star}_{1:k}\in\mathbb{R}^{k\times d_e}$.
Since the dimensionalities differ, we propose a projector
$P_\varphi:\mathbb{R}^{d_e}\to\mathbb{R}^{d_t}$ and define
\begin{equation}
u_{1:k}=P_\varphi\!\big(h^{i^\star}_{1:k}\big). 
\end{equation}
We then compute a patch-wise similarity between the student’s projected embeddings
and the teacher’s features:
\begin{equation}
\mathcal{L}_{\mathrm{align}}
=
-\frac{1}{k}\sum_{j=1}^{k}\,
\mathrm{Sim}\!\big(u_j,\; z_j\big),
\end{equation}
This objective encourages the hidden representations from the VLA’s latent feature space
to remain aligned with the teacher’s generalized visual representations,
helping preserve perceptual consistency across tasks and environments.

\subsection{Objective}

The total loss integrates the standard autoregressive action objective
with the alignment term:
\begin{equation}
    \mathcal{L}_{\mathrm{total}}
=
\mathcal{L}_{\mathrm{VLA}}
+\lambda\,\mathcal{L}_{\mathrm{align}}, \space\space \lambda>0.
\end{equation}
Here, $\mathcal{L}_{\mathrm{VLA}}$ supervises policy learning within the current environment,
while $\mathcal{L}_{\mathrm{align}}$ acts as a regularizer that limits representational drift
away from generalized visual features.
Gradients propagate through the VLA’s visual encoder $E_{\mathrm{img}}$,
text encoder $E_{\mathrm{text}}$, and transformer backbone $B_\theta$,
while the teacher encoder $E^{\star}_{\mathrm{img}}$ remains frozen,
serving as a fixed reference to stable perceptual structure. From the Platonic viewpoint, our method maintains an semantic prior
to shared, generalized VL knowledge.
Action fine-tuning alone narrows the model’s perceptual space toward the
statistics of a specific dataset or embodiment, causing the internal
features to drift away from broad generalized representations.
The alignment loss restores this balance by enforcing consistency between
the student’s intermediate features and those of a strong, pre-trained vision model that encodes more general visual–semantic relationships.

\section{Experiments}

\subsection{Evaluation setup}

We evaluate our approach in several robotics environment based on the Simpler \citep{simpler, maniskill} using proposed VL-Think task suite (\autoref{sec:vl_think}) and adopted benchmark introduced in \citep{liu2025rl4vla}, designed to assess VLA generalization across three axes: Vision, Semantics, and Execution:

\begin{itemize}
    \item \textbf{Vision} variations alter foreground and background via dynamic textures and image-level noise, testing robustness to weak and strong visual perturbations.
    \item \textbf{Semantics} introduces unseen objects and receptacles, paraphrased instructions, and multi-object or distractor scenarios that challenge compositional reasoning.
    \item \textbf{Execution} changes low-level control conditions through randomized initial poses and mid-episode object repositioning, probing action-level robustness.
\end{itemize}

OOD evaluation holds out at least one variation factor per axis, including 9 novel objects, 16 unseen receptacles, 5 new scene textures, and 16 distractor backgrounds. Additionally, we perform linear probing on ImageNet-100 \citep{tian2020goodviews} to quantify the quality of VLA's representations learned using different methods.

Each model variant is evaluated over 128 randomized seeds, we report success as mean $\pm$ standard deviation (SD). In \autoref{sec:ablations} we use the paired Wilcoxon signed-rank test \citep{wilcoxon1945individual} with one-sided alternative, and report p-values. All models are trained for the same number of epochs with identical hyperparameters to ensure fair comparison.

\subsection{Training setup}

For supervised fine-tuning, we collect 1400 expert demonstration trajectories using the MPLib motion planner \citep{motion_planner}. Training randomization spans 16 tables, 16 objects (yielding on average $\sim$5 episodes per training variation), and multiple pose perturbations. During all fine-tuning runs, LoRA adapters \citep{lora} are applied to all linear layers of the VLA.

\subsection{Baselines}

Using a widely adopted open-source OpenVLA model, we compare our proposed alignment method against several fine-tuning baselines.  

\begin{itemize}
    \item \textbf{Default} -- Standard supervised fine-tuning (SFT) using cross-entropy loss on demonstration data, serving as the primary baseline.
    \item \textbf{Freeze} -- SFT with the VLA's visual encoder weights frozen during training; this setup tests the hypothesis that frozen representations might help with generalization.
    \item \textbf{Align (ours)} -- SFT combined with our auxiliary visual representation alignment loss, described in \autoref{sec:visual_aligment}, which explicitly anchors the VLA’s vision encoder to a pretrained generalist vision teacher.
\end{itemize}

\begin{table*}[t] 
\centering
\small
\caption{VL-Think VLM results across eight domains. The benchmark reveals a strong correlation between VL understanding and model scale: larger VLMs achieve higher overall success. However, OpenVLA–7B fine-tuned for action shows clear VL degradation: its performance drops markedly compared to the original PrismaticVLM across all domains except color, where VL skills remain largely preserved.}
\label{tab:vlthink_vlm}
\begin{tabular}{lcccccccc}
\toprule
\textbf{Model} &
\textbf{Arrow} &
\textbf{Color} &
\textbf{Laundry} &
\textbf{Parity} &
\textbf{PublicInfo} &
\textbf{Shape} &
\textbf{Traffic} &
\textbf{Weather} \\
\midrule
InternVL3.5-4B    & 0.80 $\pm$ 0.02 & 0.94 $\pm$ 0.01 & 0.23 $\pm$ 0.02 & 0.54 $\pm$ 0.03 & 0.72 $\pm$ 0.03 & 0.80 $\pm$ 0.02 & 0.62 $\pm$ 0.03 & 0.75 $\pm$ 0.03 \\
InternVL3.5-8B    & 0.67 $\pm$ 0.03 & 0.94 $\pm$ 0.01 & 0.13 $\pm$ 0.02 & 0.47 $\pm$ 0.03 & 0.80 $\pm$ 0.02 & 0.77 $\pm$ 0.02 & 0.60 $\pm$ 0.03 & 0.80 $\pm$ 0.02 \\
Ovis2.5-2B        & 0.84 $\pm$ 0.02 & \textbf{0.99 $\pm$ 0.01} & 0.47 $\pm$ 0.03 & 0.55 $\pm$ 0.03 & 0.78 $\pm$ 0.02 & \textbf{0.89 $\pm$ 0.02} & 0.72 $\pm$ 0.03 & 0.88 $\pm$ 0.02 \\
Ovis2.5-9B        & \textbf{0.93 $\pm$ 0.02} & 0.94 $\pm$ 0.01 & \textbf{0.52 $\pm$ 0.03} & 0.55 $\pm$ 0.03 & \textbf{0.89 $\pm$ 0.02} & 0.87 $\pm$ 0.02 & \textbf{0.79 $\pm$ 0.02} & \textbf{0.98 $\pm$ 0.01} \\
Qwen2.5-7B        & 0.66 $\pm$ 0.03 & 0.87 $\pm$ 0.02 & 0.26 $\pm$ 0.03 & \textbf{0.58 $\pm$ 0.03} & 0.48 $\pm$ 0.03 & 0.81 $\pm$ 0.02 & 0.61 $\pm$ 0.03 & 0.70 $\pm$ 0.03 \\
Prismatic-DS-7B   & 0.47 $\pm$ 0.03 & 0.69 $\pm$ 0.03 & 0.37 $\pm$ 0.03 & 0.45 $\pm$ 0.03 & 0.62 $\pm$ 0.03 & 0.59 $\pm$ 0.03 & 0.48 $\pm$ 0.03 & 0.62 $\pm$ 0.03 \\
\midrule
OpenVLA-7B       &
0.26 $\pm$ 0.02 &
0.69 $\pm$ 0.02 &
0.30 $\pm$ 0.03 &
0.43 $\pm$ 0.02 &
0.24 $\pm$ 0.02 & 
0.40 $\pm$ 0.02 &
0.29 $\pm$ 0.02 &
0.32 $\pm$ 0.03 \\
OpenVLA-7B Align  &
0.24 $\pm$ 0.02 &
\textbf{0.82 $\pm$ 0.02} &
0.29 $\pm$ 0.03 &
0.42  $\pm$ 0.03 &
\textbf{0.30 $\pm$ 0.03} &
\textbf{0.48 $\pm$ 0.02} &
0.28 $\pm$ 0.03 &
0.27 $\pm$ 0.02 \\
\bottomrule
\end{tabular}
\end{table*}

\begin{table}[b]
\centering
\caption{Linear probing results. OpenVLA Align retains stronger features than both the pretrained and SFT variants, closing much of the gap to the C-RADIOv3 teacher and demonstrating improved semantic consistency after action fine-tuning.}
\label{tab:linear_probing}

\begingroup
\renewcommand{\arraystretch}{0.95}
\setlength{\aboverulesep}{0.3ex}
\setlength{\belowrulesep}{0.3ex}
\setlength{\tabcolsep}{0pt}

\makebox[\linewidth][c]{%
\begin{tabular}{@{\hspace{6mm}} l @{\hspace{16mm}} c @{\hspace{6mm}}}
\toprule
\textbf{Model} & \textbf{Accuracy (\%)} \\
\midrule
C-RADIOv3          & $\boldsymbol{87.31}$ \\
OpenVLA Align      & $\underline{82.13}$ \\
OpenVLA Pretrained & $79.88$ \\
OpenVLA SFT        & $77.48$ \\
\bottomrule
\end{tabular}%
}
\endgroup
\end{table}

\subsection{Results: OOD Evaluation}

Results in \autoref{tab:ood_performance} shows that our visual alignment method yields consistent improvements across all evaluation axes: Semantic, Vision, and Execution. This result underscores the effectiveness of visual representation alignment in enhancing robustness to visual shifts, text instruction variations, texture changes, and background perturbations that frequently occur in real-world scenarios. The improvement indicates that aligning internal visual-language embeddings not only stabilizes perception but also reinforces the semantic grounding. Conversely, the Freeze baseline completely fails across all categories (as also observed in \citep{wang2025vlaadaptereffectiveparadigmtinyscale}), yielding near-zero performance. This confirms that simply freezing the pretrained visual encoder does not preserve useful representations during adaptation. Without joint optimization, the frozen features become mismatched with the evolving action components, leading to severe degradation of both perception and control. 

Overall, these results validate that visual alignment serves as an effective regularizer against representation degradation, allowing the model to recover general-purpose visual semantics while adapting to new robotic environments.

\subsection{Results: Linear probing}
\label{sec:res_lin_prob}

To further evaluate the representational quality learned by our model, we conduct a linear probing analysis on the ImageNet-100 dataset~\citep{tian2020goodviews}. Specifically, we extract patch embeddings from the final layer of the C-RADIOv3~\citep{radio} teacher and from the intermediate visual layers of different OpenVLA variants. Following standard practice in representation learning~\citep{repa, platonicrepresentationhypothesis}, we freeze the visual encoders and train a single linear classifier on top of their frozen features to measure the separability of semantic categories. This setup directly quantifies how linearly decodable the visual features remain after action fine-tuning. 

The results summarized in \autoref{tab:linear_probing} reveal several consistent trends. As expected, the C-RADIOv3 teacher achieves the highest probing accuracy, reflecting its strong pretrained representations. Among the VLA variants, the OpenVLA fine-tuned with our proposed Visual Representation Alignment method outperforms both the pretrained checkpoint and the model fine-tuned with naive SFT. This improvement indicates that our alignment strategy effectively enhances the VLA's representations during action fine-tuning. In contrast, naive SFT substantially reduces probing accuracy relative to the pretrained model, confirming that standard fine-tuning harms representational quality. Our aligned model not only mitigates this degradation but surpasses the pretrained baseline, indicating that the alignment loss strengthens semantic consistency and leads to more transferable visual features.

\subsection{Results: VL-Think}

Following the experiments in \autoref{sec:vl_think_eval1}, we evaluate of OpenVLA fine-tuned with our proposed visual representation alignment method (OpenVLA-7B Align), under identical data, budget, and evaluation settings. Results in \autoref{tab:vlthink_vlm} show that \textbf{SFT-Align} partially mitigates domain forgetting observed under default SFT. In particular, performance on \textit{Color} and \textit{Shape} domains consistently improves even surpassing the PrismaticVLM upper bound, but leaving other domains mostly unchanged.

These outcomes highlight both the promise and limits of the proposed representation alignment under constrained settings. We hypothesize that the modest size and diversity of the SFT dataset and the limited expressivity of LoRA updates are insufficient to restore less frequent VL concepts that are underrepresented in robotics data. We hypothesize that expanding data breadth and relaxing parameter-efficiency constraints will unlock broader gains beyond commonly represented domains. Verifying this hypothesis is an important direction for future work.

\section{Ablations}
\label{sec:ablations}

In this section, we conduct a systematic ablation study to analyze how different design choices affect the performance of our visual alignment method. We examine the impact of the teacher model used for alignment, the alignment strategy and target layers, the projector type and the loss functions. Together, these experiments provide insights into which components are most critical for effective alignment of visual representations.

\subsection{Visual teacher models}
A key question in our approach concerns the choice of the teacher model that provides reference representations for alignment. 
From the Platonic perspective, each vision foundation encoder captures a different projection of broadly generalizable visual knowledge, 
and alignment to a stronger teacher helps preserve these high-level, transferable abstractions within the VLA during fine-tuning. 
We therefore examine whether foundation models trained on large-scale, diverse, and multi-view data yield better alignment and stronger transfer.

\begin{table}[b]
\centering
\caption{Comparison of pretrained Teacher Vision Models across generalization dimensions. 
Values represent mean $\pm$ aggregated across all environments within each dimension and p-value. 
Best results per column are highlighted in bold (for more details see \autoref{tab:app_abb_vision_teacher} from Appendix).}
\setlength{\tabcolsep}{6pt}
\renewcommand{\arraystretch}{1.15}
\resizebox{0.47\textwidth}{!}{%
\begin{tabular}{lccc}
\toprule
\textbf{Teacher} & \textbf{Semantic} & \textbf{Vision} & \textbf{Execution} \\
\midrule
C-RADIOv3 & $\boldsymbol{0.61}$ & $\boldsymbol{0.72}$ & $\boldsymbol{0.39}$ \\
DINOv2 & $0.57$ \small{(p=0.05)} & $\underline{0.69}$ \small{(p=0.12)} & $\underline{0.37}$ \small{(p=0.43)} \\
SigLIP & $0.54$ \small{(p=0.01)} & $0.65$ \small{(p=0.03)} & $0.35$ \small{(p=0.09)} \\
Theia & $0.56$ \small{(p=0.03)} & $0.67$ \small{(p=0.05)} & $0.36$ \small{(p=0.15)} \\
\bottomrule
\label{tab:teacher_backbone}
\end{tabular}%
}
\end{table}

To test this, we evaluate several state-of-the-art vision encoders, including DINOv2~\citep{dinov2}, SigLIP~\citep{siglip}, C-RADIOv3~\citep{radio}, and Theia~\citep{theia}. As shown in \autoref{tab:teacher_backbone}, C-RADIOv3 achieves the best overall results, 
indicating that stronger and more capable vision models those trained on large-scale, semantically rich, and multimodal data
offer more stable and generalizable visual features for alignment. Such teachers serve as stronger Platonic anchors, guiding the VLA to align with transferable and semantically consistent representations that improve robustness across tasks and domains.

\subsection{Alignment method}

\begin{table}[t]
\centering
\caption{Comparison of alignment paradigms across generalization dimensions. Reported as mean across dimensions and p-value, best results are highlighted in bold.}
\setlength{\tabcolsep}{6pt}
\renewcommand{\arraystretch}{1.15}
\resizebox{0.45\textwidth}{!}{
\begin{tabular}{lccc}
\toprule
\textbf{Method} & \textbf{Semantic} & \textbf{Vision} & \textbf{Execution} \\
\midrule
Backbone2Enc & $\boldsymbol{0.61}$ & $\boldsymbol{0.72}$ & $\boldsymbol{0.39}$ \\
Enc2Enc & $0.55$ \small{(p=0.01)} & $0.66$ \small{(p=0.04)} & $\underline{0.38}$ \small{(p=0.64)} \\
\bottomrule
\label{tab:align_method}
\end{tabular}
}
\end{table}

We next evaluate different alignment paradigms to determine which level of the VLA model benefits most from visual representation alignment. Two principal strategies are tested:

\begin{enumerate}
    \item \textbf{Backbone2Enc} -- Aligning the representations of the VLA's transformer backbone to the final-layer features of the teacher’s visual encoder.
    \item \textbf{Enc2Enc} -- Aligning the features of the VLA’s own visual encoder directly to the teacher model’s final embeddings.
\end{enumerate}

Our experiments reveal (\autoref{tab:align_method}) that \textit{Backbone2Enc} consistently yields stronger results. This indicates that the primary representational degradation occurs not in the early encoder layers but in the middle-to-late fusion layers, where VL integration and task-specific adaptation are most active. Regularizing these deeper representations appears crucial for maintaining visual–semantic consistency while allowing the lower layers to adapt freely to domain-specific low-level cues.

\subsection{Projector type}
\label{sec:proj_type}

\begin{table}[b]
\centering
\caption{Comparison of different projection methods across generalization dimensions (mean across dimensions, p-value). Each projection type was evaluated in both frozen and trainable variants (for detailed results see \autoref{tab:app_abb_vision_projector} from Appendix).}
\setlength{\tabcolsep}{6pt}
\renewcommand{\arraystretch}{1.15}
\resizebox{0.45\textwidth}{!}{%
\begin{tabular}{lcccc}
\toprule
\textbf{Projector} & \textbf{Freeze} & \textbf{Semantic} & \textbf{Vision} & \textbf{Execution} \\
\midrule
MLP & \cmark & $\boldsymbol{0.61}$ & $\boldsymbol{0.72}$ & $\boldsymbol{0.39}$ \\
MLP & \xmark & $0.54$ \small{(p<0.01)} & $0.71$ \small{(p=0.48)} & $0.32$ \small{(p=0.06)} \\
Cosine & \xmark & $0.59$ \small{(p=0.08)} & $0.71$ \small{(p=0.13)} & $0.38$ \small{(p=0.45)} \\
OrthProj & \cmark & $0.55$ \small{(p=0.01)} & $0.71$ \small{(p=0.37)} & $0.38$ \small{(p=0.45)} \\
FILM & \xmark & $0.54$ \small{(p<0.01)} & $0.69$ \small{(p=0.11)} & $0.35$ \small{(p=0.18)} \\
Whitening & \xmark & $0.56$ \small{(p=0.01)} & $\underline{0.72}$ \small{(p=0.61)} & $\underline{0.44}$ \small{(p=0.97)}\\
Spectral & \cmark & $\underline{0.58}$ \small{(p=0.17)} & $0.71$ \small{(p=0.26)} & $0.39$ \small{(p=0.65)} \\
\bottomrule
\label{tab:proj_type}
\end{tabular}%
}
\end{table}

To evaluate how different projection mappings affect representation alignment, we compare several projector variants that map the VLA’s hidden states $\mathbb{R}^{d_{\text{e}}}$ to the teacher’s embedding space $\mathbb{R}^{d_t}$. All projectors share identical input–output dimensions but differ in their internal transformation $P_{\varphi}: \mathbb{R}^{d_{\text{e}}} \rightarrow \mathbb{R}^{d_t}$.

We examine multiple projection strategies, including linear, cosine-similarity–based, orthogonal, spectral-normalized, FiLM-conditioned, Whitening–affine, and MLP-based mappings. Our experiments show that the frozen MLP projector yields the most reliable, robust alignment across all evaluation dimensions. We hypothesize that freezing the projector is critical in our setup: when trainable, the model minimizes alignment loss primarily through projector adaptation rather than meaningful changes in the VLA’s internal representations. In this case, the projector quickly learns to output embeddings that merely approximate the teacher’s space, effectively bypassing representational correction. We attribute this to two factors: the relatively small amount of alignment data and the substantial dimensionality gap between the vision teacher and the VLA backbone embeddings ($d_t=768$, $d_e=4096$). Freezing the projector constrains this shortcut, forcing the alignment objective to act directly on the student’s hidden representations, yielding more semantically grounded and transferable feature alignment.

\subsection{Alignment layers}

\begin{table}[H]
\centering
\caption{Comparison of different layers for alignment across generalization dimensions  (mean across dimensions, p-value) (for detailed results see \autoref{tab:app_layers} from Appendix).}
\setlength{\tabcolsep}{6pt}
\renewcommand{\arraystretch}{1.15}
\resizebox{0.45\textwidth}{!}{%
\begin{tabular}{lccc}
\toprule
\textbf{Method} & \textbf{Semantic} & \textbf{Vision} & \textbf{Execution} \\
\midrule
Middle & $\boldsymbol{0.61}$ & $\boldsymbol{0.72}$ & $0.39$ \\
Early  & $0.51$ \small{(p<0.01)} & $0.66$ \small{(p=0.04)} & $\underline{0.38}$ \small{(p=0.85)} \\
Late & $0.54$ \small{(p=0.03)} & $\underline{0.69}$ \small{(p=0.83)} & $0.36$ \small{(p=0.52)} \\
\bottomrule
\label{tab:alig_layers}
\end{tabular}%
}
\end{table}

We further investigate which layers within the VLA transformer’s backbone should be aligned to achieve the most effective representation recovery. Prior literature on VLM interpretability \citep{zhang2025mllms} and our own analyses (\autoref{sec:motivation}) suggest that middle layers are primarily responsible for VL fusion and semantic grounding, whereas early layers encode low-level features and later layers specialize in action prediction. Accordingly, we perform experiments aligning different types of layers: Early, Middle, Late. The results (\autoref{tab:alig_layers}) confirm that the middle layers play a central role in semantic grounding and and aligning them yields the most substantial improvements across generalization axes.

\subsection{Loss functions and alignment coefficient}
\begin{table}[H]
\centering
\caption{Comparison of different loss functions across generalization dimensions (mean across dimensions, p-value).}
\setlength{\tabcolsep}{6pt}
\renewcommand{\arraystretch}{1.15}
\resizebox{0.45\textwidth}{!}{%
\begin{tabular}{lccc}
\toprule
\textbf{Objective} & \textbf{Semantic} & \textbf{Vision} & \textbf{Execution} \\
\midrule
Cosine & $\boldsymbol{0.61}$ & $\boldsymbol{0.72}$ & $\boldsymbol{0.39}$ \\
L2& $0.54$ \small{(p<0.01)} & $0.63$ \small{(p<0.01)} & $0.34$ \small{(p=0.05)}\\
InfoNCE & $0.57$ \small{(p=0.05)} & $0.64$ \small{(p=0.04)} & $\underline{0.36}$ \small{(p=0.21)}\\
\bottomrule
\label{tab:agg_loss_func}
\end{tabular}%
}
\end{table}

Finally, we assess the impact of the alignment loss and its weighting coefficient. We test several variants, including cosine similarity (Cossim), L2, and contrastive NT-Xent \citep{chen2020simpleframeworkcontrastivelearning} losses, across alignment coefficients
$\lambda$ = \{0.2, 0.5, 1.0, 3.0\}. The results demonstrate (\autoref{tab:agg_loss_func}) that Cossim loss achieves the most stable and consistent improvements, particularly when the auxiliary weight is set to $\lambda$ = 0.2. This setting effectively constrains representation drift without overpowering the task objective.

\section{Conclusion}
In this work, we examined how fine-tuning VLA models on robotic tasks leads to degradation of VL understanding and representation quality. To analyze this effect, we introduced the VL-Think diagnostic suite and interpretability probes, including attention map analyses and linear probing, which reveal how VL skills degrade during action fine-tuning. To address this issue, we proposed a lightweight Visual Alignment method that anchors the VLA to its pretrained visual teacher, consistently improving OOD generalization across diverse domains including novel objects, unseen scene compositions, texture and lighting variations, and instruction paraphrases. Due to compute constraints, our study focused on fine-tuning rather than full-scale pretraining. We hope this study guides future efforts toward scalable robotic pretraining and systematic evaluation of how VLAs inherit and retain VL knowledge from VLMs.

\bibliographystyle{ACM-Reference-Format} 
\bibliography{bibtex}

\clearpage

\appendix

\section{Appendix}

This appendix provides additional technical details, extended results, and supplementary materials that support and complement the main findings presented in the paper. We include comprehensive ablations on alignment strategies, projector architectures, teacher models, and layer selection and alignment coefficient. These materials aim to enhance the transparency, reproducibility, and interpretability of our proposed method.

\subsection{Training Hyperparameters}
\autoref{tab:openvla_train_config_alig} presents the training parameters used during the visual alignment fine-tuning. All other training parameters remained unchanged across both experiments and model variants.

\begin{table}[h]
\centering
\caption{Best training configuration for OpenVLA fine-tuning with visual alignment. 
All other methods were trained with identical hyperparameters except for the alignment-specific settings.
}
\setlength{\tabcolsep}{6pt}
\renewcommand{\arraystretch}{1.15}
\resizebox{0.38\textwidth}{!}{%
\begin{tabular}{lc}
\toprule
\textbf{Parameter} & \textbf{Value} \\
\midrule
Fine-tuning steps & 60000 \\
Batch size & 8 \\
Gradient accumulation steps & 1 \\
Learning rate & $5\times10^{-4}$ \\
LoRA rank & 32 \\
Alignment coefficient & 0.2 \\
Alignment projector & MLP Ln\&D \\
Alignment method & Backbone2Enc \\
Aligned layers & 16 \\
Mode & \texttt{alig} \\
Projector dimension & 2048 \\
Freeze alignment projector & \cmark \\
\bottomrule
\label{tab:openvla_train_config_alig}
\end{tabular}%
}
\end{table}

\begin{table*}[t]
\centering
\caption{Comparison on projection methods, table shows performance across environments (mean $\pm$ SD).}
\setlength{\tabcolsep}{3pt}
\renewcommand{\arraystretch}{1.15}
\resizebox{\textwidth}{!}{%
\begin{tabular}{lccccccccccccccc}
\toprule
\textbf{Method} & \textbf{Freeze} &
\multicolumn{5}{c}{\textbf{Semantic}} &
\multicolumn{5}{c}{\textbf{Vision}} &
\multicolumn{3}{c}{\textbf{Execution}} \\
\cmidrule(lr){3-7}
\cmidrule(lr){8-12}
\cmidrule(lr){13-15}
& & Carrot & Instruct & MultiCarrot & MultiPlate & Plate
& VisionImg & Tex03 & Tex05 & Whole03 & Whole05
& Position & EEPose & PosChangeTo \\
\midrule
\textbf{MLP} & \xmark &
$0.49{\scriptsize \pm 0.05}$ &
$0.79{\scriptsize \pm 0.02}$ &
$0.27{\scriptsize \pm 0.01}$ &
$0.46{\scriptsize \pm 0.02}$ &
$0.72{\scriptsize \pm 0.01}$ &
$0.85{\scriptsize \pm 0.01}$ &
$0.73{\scriptsize \pm 0.02}$ &
$0.58{\scriptsize \pm 0.03}$ &
$0.77{\scriptsize \pm 0.03}$ &
$\underline{0.66{\scriptsize \pm 0.02}}$ &
$0.53{\scriptsize \pm 0.05}$ &
$0.31{\scriptsize \pm 0.01}$ &
$0.21{\scriptsize \pm 0.04}$ \\
\textbf{MLP} & \cmark &
$\boldsymbol{0.61{\scriptsize \pm 0.01}}$ &
$\underline{0.83{\scriptsize \pm 0.03}}$ &
$\boldsymbol{0.35{\scriptsize \pm 0.02}}$ &
$0.49{\scriptsize \pm 0.02}$ &
$\boldsymbol{0.75{\scriptsize \pm 0.01}}$ &
$\underline{0.86{\scriptsize \pm 0.02}}$ &
$0.70{\scriptsize \pm 0.02}$ &
$\boldsymbol{0.67{\scriptsize \pm 0.02}}$ &
$\boldsymbol{0.80{\scriptsize \pm 0.02}}$ &
$0.60{\scriptsize \pm 0.02}$ &
$\underline{0.58{\scriptsize \pm 0.02}}$ &
$\underline{0.38{\scriptsize \pm 0.02}}$ &
$0.20{\scriptsize \pm 0.03}$ \\
\textbf{Cosine} & \cmark &
$0.49{\scriptsize \pm 0.04}$ &
$0.82{\scriptsize \pm 0.03}$ &
$0.30{\scriptsize \pm 0.06}$ &
$0.45{\scriptsize \pm 0.05}$ &
$0.73{\scriptsize \pm 0.03}$ &
$0.75{\scriptsize \pm 0.03}$ &
$0.71{\scriptsize \pm 0.04}$ &
$0.54{\scriptsize \pm 0.05}$ &
$0.77{\scriptsize \pm 0.04}$ &
$0.60{\scriptsize \pm 0.04}$ &
$0.51{\scriptsize \pm 0.01}$ &
$0.32{\scriptsize \pm 0.05}$ &
$0.29{\scriptsize \pm 0.07}$ \\
\textbf{Cosine} & \xmark &
$0.53{\scriptsize \pm 0.02}$ &
$0.79{\scriptsize \pm 0.01}$ &
$\underline{0.34{\scriptsize \pm 0.04}}$ &
$\boldsymbol{0.55{\scriptsize \pm 0.02}}$ &
$0.69{\scriptsize \pm 0.05}$ &
$0.80{\scriptsize \pm 0.01}$ &
$0.73{\scriptsize \pm 0.03}$ &
$0.60{\scriptsize \pm 0.03}$ &
$0.75{\scriptsize \pm 0.03}$ &
$0.61{\scriptsize \pm 0.03}$ &
$0.55{\scriptsize \pm 0.09}$ &
$0.35{\scriptsize \pm 0.03}$ &
$0.25{\scriptsize \pm 0.04}$ \\
\textbf{FILM} & \xmark &
$0.53{\scriptsize \pm 0.04}$ &
$\boldsymbol{0.84{\scriptsize \pm 0.03}}$ &
$0.35{\scriptsize \pm 0.04}$ &
$0.44{\scriptsize \pm 0.03}$ &
$0.68{\scriptsize \pm 0.01}$ &
$0.83{\scriptsize \pm 0.03}$ &
$0.72{\scriptsize \pm 0.04}$ &
$0.61{\scriptsize \pm 0.02}$ &
$0.75{\scriptsize \pm 0.06}$ &
$0.60{\scriptsize \pm 0.07}$ &
$0.58{\scriptsize \pm 0.09}$ &
$0.31{\scriptsize \pm 0.03}$ &
$0.26{\scriptsize \pm 0.03}$ \\
\textbf{FILM} & \cmark &
$0.52{\scriptsize \pm 0.02}$ &
$0.77{\scriptsize \pm 0.02}$ &
$0.34{\scriptsize \pm 0.03}$ &
$0.43{\scriptsize \pm 0.05}$ &
$0.67{\scriptsize \pm 0.03}$ &
$0.79{\scriptsize \pm 0.03}$ &
$0.75{\scriptsize \pm 0.03}$ &
$0.53{\scriptsize \pm 0.04}$ &
$0.76{\scriptsize \pm 0.01}$ &
$0.64{\scriptsize \pm 0.04}$ &
$0.57{\scriptsize \pm 0.02}$ &
$0.27{\scriptsize \pm 0.03}$ &
$0.24{\scriptsize \pm 0.02}$ \\

\textbf{OrthProj} & \cmark &
$0.54{\scriptsize \pm 0.01}$ &
$0.78{\scriptsize \pm 0.03}$ &
$0.31{\scriptsize \pm 0.03}$ &
$0.44{\scriptsize \pm 0.06}$ &
$0.71{\scriptsize \pm 0.04}$ &
$0.84{\scriptsize \pm 0.02}$ &
$0.73{\scriptsize \pm 0.01}$ &
$0.58{\scriptsize \pm 0.01}$ &
$0.72{\scriptsize \pm 0.05}$ &
$0.60{\scriptsize \pm 0.03}$ &
$0.57{\scriptsize \pm 0.04}$ &
$0.29{\scriptsize \pm 0.02}$ &
$0.20{\scriptsize \pm 0.02}$ \\
\textbf{OrthProj} & \xmark &
$0.49{\scriptsize \pm 0.04}$ &
$0.77{\scriptsize \pm 0.02}$ &
$0.33{\scriptsize \pm 0.02}$ &
$0.48{\scriptsize \pm 0.08}$ &
$0.70{\scriptsize \pm 0.04}$ &
$0.86{\scriptsize \pm 0.03}$ &
$\boldsymbol{0.77{\scriptsize \pm 0.01}}$ &
$0.58{\scriptsize \pm 0.05}$ &
$0.74{\scriptsize \pm 0.06}$ &
$0.63{\scriptsize \pm 0.03}$ &
$0.55{\scriptsize \pm 0.08}$ &
$0.38{\scriptsize \pm 0.03}$ &
$0.23{\scriptsize \pm 0.03}$ \\
\textbf{RFF} & \cmark &
$0.46{\scriptsize \pm 0.02}$ &
$0.74{\scriptsize \pm 0.03}$ &
$0.31{\scriptsize \pm 0.03}$ &
$0.43{\scriptsize \pm 0.04}$ &
$0.70{\scriptsize \pm 0.02}$ &
$0.84{\scriptsize \pm 0.04}$ &
$0.72{\scriptsize \pm 0.03}$ &
$0.61{\scriptsize \pm 0.03}$ &
$0.75{\scriptsize \pm 0.05}$ &
$0.56{\scriptsize \pm 0.04}$ &
$0.56{\scriptsize \pm 0.02}$ &
$0.32{\scriptsize \pm 0.04}$ &
$0.20{\scriptsize \pm 0.03}$ \\
\textbf{RFF} & \xmark &
$0.56{\scriptsize \pm 0.03}$ &
$0.80{\scriptsize \pm 0.02}$ &
$0.30{\scriptsize \pm 0.04}$ &
$0.48{\scriptsize \pm 0.02}$ &
$0.70{\scriptsize \pm 0.02}$ &
$0.78{\scriptsize \pm 0.02}$ &
$0.72{\scriptsize \pm 0.02}$ &
$0.56{\scriptsize \pm 0.02}$ &
$0.72{\scriptsize \pm 0.03}$ &
$0.62{\scriptsize \pm 0.02}$ &
$0.57{\scriptsize \pm 0.01}$ &
$0.27{\scriptsize \pm 0.03}$ &
$0.27{\scriptsize \pm 0.04}$ \\
\textbf{Whitening} & \cmark &
$0.46{\scriptsize \pm 0.03}$ &
$0.74{\scriptsize \pm 0.04}$ &
$0.21{\scriptsize \pm 0.02}$ &
$0.33{\scriptsize \pm 0.02}$ &
$0.64{\scriptsize \pm 0.02}$ &
$0.80{\scriptsize \pm 0.05}$ &
$0.72{\scriptsize \pm 0.03}$ &
$0.57{\scriptsize \pm 0.04}$ &
$0.71{\scriptsize \pm 0.02}$ &
$0.57{\scriptsize \pm 0.04}$ &
$0.41{\scriptsize \pm 0.03}$ &
$0.18{\scriptsize \pm 0.01}$ &
$0.12{\scriptsize \pm 0.01}$ \\
\textbf{Whitening} & \xmark &
$0.52{\scriptsize \pm 0.03}$ &
$0.79{\scriptsize \pm 0.02}$ &
$0.32{\scriptsize \pm 0.02}$ &
$0.48{\scriptsize \pm 0.02}$ &
$0.72{\scriptsize \pm 0.02}$ &
$0.81{\scriptsize \pm 0.03}$ &
$\underline{0.75{\scriptsize \pm 0.05}}$ &
$\underline{0.64{\scriptsize \pm 0.04}}$ &
$\underline{0.77{\scriptsize \pm 0.01}}$ &
$\boldsymbol{0.68{\scriptsize \pm 0.01}}$ &
$\boldsymbol{0.63{\scriptsize \pm 0.03}}$ &
$\boldsymbol{0.39{\scriptsize \pm 0.02}}$ &
$\boldsymbol{0.32{\scriptsize \pm 0.05}}$ \\
\textbf{Spectral} & \cmark &
$0.53{\scriptsize \pm 0.02}$ &
$0.81{\scriptsize \pm 0.03}$ &
$0.34{\scriptsize \pm 0.07}$ &
$\underline{0.51{\scriptsize \pm 0.04}}$ &
$\underline{0.74{\scriptsize \pm 0.05}}$ &
$\boldsymbol{0.91{\scriptsize \pm 0.03}}$ &
$0.73{\scriptsize \pm 0.03}$ &
$0.54{\scriptsize \pm 0.06}$ &
$0.75{\scriptsize \pm 0.03}$ &
$0.61{\scriptsize \pm 0.03}$ &
$0.64{\scriptsize \pm 0.04}$ &
$0.32{\scriptsize \pm 0.03}$ &
$0.23{\scriptsize \pm 0.04}$ \\
\bottomrule
\textbf{Default} & - &
$0.49{\scriptsize \pm 0.02}$ &
$0.74{\scriptsize \pm 0.02}$ &
$0.28{\scriptsize \pm 0.02}$ &
$0.43{\scriptsize \pm 0.02}$ &
$0.73{\scriptsize \pm 0.02}$ &
$0.81{\scriptsize \pm 0.01}$ &
$0.67{\scriptsize \pm 0.01}$ &
$0.55{\scriptsize \pm 0.03}$ &
$0.71{\scriptsize \pm 0.02}$ &
$0.56{\scriptsize \pm 0.01}$ &
$0.43{\scriptsize \pm 0.02}$ &
$0.34{\scriptsize \pm 0.01}$ &
$0.23{\scriptsize \pm 0.01}$ \\
\bottomrule
\label{tab:app_abb_vision_projector}
\end{tabular}%
}
\end{table*}

\begin{table*}[t]
\centering
\caption{Comparison on vision teacher models, table shows performance across environments (mean $\pm$ SD).}
\setlength{\tabcolsep}{3pt}
\renewcommand{\arraystretch}{1.15}
\resizebox{\textwidth}{!}{%
\begin{tabular}{lccccccccccccc}
\toprule
\textbf{Teacher} &
\multicolumn{5}{c}{\textbf{Semantic}} &
\multicolumn{5}{c}{\textbf{Vision}} &
\multicolumn{3}{c}{\textbf{Execution}} \\
\cmidrule(lr){2-6}
\cmidrule(lr){7-11}
\cmidrule(lr){12-14}
& Carrot & Instruct & MultiCarrot & MultiPlate & Plate
& VisionImg & Tex03 & Tex05 & Whole03 & Whole05
& Position & EEPose & PosChangeTo \\
\midrule
\textbf{C-RADIOv3-ViT-L} &
$\boldsymbol{0.61{\scriptsize \pm 0.01}}$ &
$0.83{\scriptsize \pm 0.03}$ &
$\boldsymbol{0.35{\scriptsize \pm 0.02}}$ &
$\boldsymbol{0.49{\scriptsize \pm 0.02}}$ &
$\boldsymbol{0.75{\scriptsize \pm 0.01}}$ &
$\boldsymbol{0.86{\scriptsize \pm 0.02}}$ &
$0.70{\scriptsize \pm 0.02}$ &
$\boldsymbol{0.67{\scriptsize \pm 0.02}}$ &
$\boldsymbol{0.80{\scriptsize \pm 0.02}}$ &
$0.60{\scriptsize \pm 0.02}$ &
$\underline{0.58{\scriptsize \pm 0.02}}$ &
$\underline{0.38{\scriptsize \pm 0.02}}$ &
$0.20{\scriptsize \pm 0.03}$ \\
\textbf{DINOv2-ViT-L} &
$0.49{\scriptsize \pm 0.02}$ &
$0.74{\scriptsize \pm 0.04}$ &
$0.31{\scriptsize \pm 0.02}$ &
$0.46{\scriptsize \pm 0.03}$ &
$0.73{\scriptsize \pm 0.03}$ &
$0.80{\scriptsize \pm 0.01}$ &
$0.72{\scriptsize \pm 0.04}$ &
$0.57{\scriptsize \pm 0.04}$ &
$0.72{\scriptsize \pm 0.01}$ &
$\boldsymbol{0.65{\scriptsize \pm 0.02}}$ &
$0.55{\scriptsize \pm 0.03}$ &
$0.33{\scriptsize \pm 0.02}$ &
$0.21{\scriptsize \pm 0.03}$ \\
\textbf{DINOv2-ViT-G} &
$0.55{\scriptsize \pm 0.04}$ &
$\boldsymbol{0.84{\scriptsize \pm 0.01}}$ &
$0.32{\scriptsize \pm 0.05}$ &
$0.42{\scriptsize \pm 0.03}$ &
$\underline{0.74{\scriptsize \pm 0.01}}$ &
$0.83{\scriptsize \pm 0.02}$ &
$\boldsymbol{0.72{\scriptsize \pm 0.01}}$ &
$0.60{\scriptsize \pm 0.03}$ &
$0.72{\scriptsize \pm 0.02}$ &
$0.58{\scriptsize \pm 0.02}$ &
$\boldsymbol{0.58{\scriptsize \pm 0.04}}$ &
$0.34{\scriptsize \pm 0.02}$ &
$0.20{\scriptsize \pm 0.04}$ \\
\textbf{Theia} &
$0.52{\scriptsize \pm 0.03}$ &
$0.76{\scriptsize \pm 0.04}$ &
$0.29{\scriptsize \pm 0.03}$ &
$0.39{\scriptsize \pm 0.01}$ &
$0.70{\scriptsize \pm 0.01}$ &
$0.74{\scriptsize \pm 0.04}$ &
$0.70{\scriptsize \pm 0.03}$ &
$0.53{\scriptsize \pm 0.05}$ &
$0.72{\scriptsize \pm 0.08}$ &
$0.60{\scriptsize \pm 0.02}$ &
$0.53{\scriptsize \pm 0.02}$ &
$\boldsymbol{0.41{\scriptsize \pm 0.03}}$ &
$0.21{\scriptsize \pm 0.04}$ \\
\textbf{SigLIP} &
$0.36{\scriptsize \pm 0.03}$ &
$0.65{\scriptsize \pm 0.01}$ &
$0.15{\scriptsize \pm 0.02}$ &
$0.26{\scriptsize \pm 0.06}$ &
$0.57{\scriptsize \pm 0.06}$ &
$0.69{\scriptsize \pm 0.05}$ &
$0.58{\scriptsize \pm 0.05}$ &
$0.47{\scriptsize \pm 0.02}$ &
$0.64{\scriptsize \pm 0.05}$ &
$0.48{\scriptsize \pm 0.04}$ &
$0.52{\scriptsize \pm 0.04}$ &
$0.32{\scriptsize \pm 0.02}$ &
$0.18{\scriptsize \pm 0.04}$ \\
\bottomrule
\textbf{Default} &
$0.49{\scriptsize \pm 0.02}$ &
$0.74{\scriptsize \pm 0.02}$ &
$0.28{\scriptsize \pm 0.02}$ &
$0.43{\scriptsize \pm 0.02}$ &
$0.73{\scriptsize \pm 0.02}$ &
$0.81{\scriptsize \pm 0.01}$ &
$0.67{\scriptsize \pm 0.01}$ &
$0.55{\scriptsize \pm 0.03}$ &
$0.71{\scriptsize \pm 0.02}$ &
$0.56{\scriptsize \pm 0.01}$ &
$0.43{\scriptsize \pm 0.02}$ &
$0.34{\scriptsize \pm 0.01}$ &
$\boldsymbol{0.23{\scriptsize \pm 0.01}}$ \\

\bottomrule
\label{tab:app_abb_vision_teacher}
\end{tabular}%
}
\end{table*}

\begin{table*}[t]
\centering
\caption{Comparison on different aligning layers, table shows performance across environments (mean $\pm$ SD).}
\setlength{\tabcolsep}{3pt}
\renewcommand{\arraystretch}{1.15}
\resizebox{\textwidth}{!}{%
\begin{tabular}{lcccccccccccccc}
\toprule
\textbf{Teacher} & \textbf{Layer (L)} &
\multicolumn{5}{c}{\textbf{Semantic}} &
\multicolumn{5}{c}{\textbf{Vision}} &
\multicolumn{3}{c}{\textbf{Execution}} \\
\cmidrule(lr){3-7}
\cmidrule(lr){8-12}
\cmidrule(lr){13-15}
& & Carrot & Instruct & MultiCarrot & MultiPlate & Plate
& VisionImg & Tex03 & Tex05 & Whole03 & Whole05
& Position & EEPose & PosChangeTo \\
\midrule
\textbf{C-RADIOv3-ViT-L} & 8 &
$0.49{\scriptsize \pm 0.02}$ &
$0.79{\scriptsize \pm 0.02}$ &
$0.27{\scriptsize \pm 0.04}$ &
$0.45{\scriptsize \pm 0.01}$ &
$0.76{\scriptsize \pm 0.04}$ &
$0.84{\scriptsize \pm 0.03}$ &
$0.72{\scriptsize \pm 0.03}$ &
$0.60{\scriptsize \pm 0.02}$ &
$0.76{\scriptsize \pm 0.04}$ &
$0.59{\scriptsize \pm 0.07}$ &
$0.59{\scriptsize \pm 0.08}$ &
$\boldsymbol{0.39{\scriptsize \pm 0.03}}$ &
$0.23{\scriptsize \pm 0.06}$ \\
\textbf{C-RADIOv3-ViT-L} & 16 &
$\boldsymbol{0.61{\scriptsize \pm 0.01}}$ &
$\boldsymbol{0.83{\scriptsize \pm 0.03}}$ &
$\boldsymbol{0.35{\scriptsize \pm 0.02}}$ &
$0.49{\scriptsize \pm 0.02}$ &
$0.75{\scriptsize \pm 0.01}$ &
$0.86{\scriptsize \pm 0.02}$ &
$0.70{\scriptsize \pm 0.02}$ &
$\boldsymbol{0.67{\scriptsize \pm 0.02}}$ &
$\boldsymbol{0.80{\scriptsize \pm 0.02}}$ &
$0.60{\scriptsize \pm 0.02}$ &
$0.58{\scriptsize \pm 0.02}$ &
$\underline{0.38{\scriptsize \pm 0.02}}$ &
$0.20{\scriptsize \pm 0.03}$ \\
\textbf{C-RADIOv3-ViT-L} & 20 &
$0.54{\scriptsize \pm 0.02}$ &
$0.81{\scriptsize \pm 0.01}$ &
$0.31{\scriptsize \pm 0.02}$ &
$0.51{\scriptsize \pm 0.02}$ &
$0.72{\scriptsize \pm 0.04}$ &
$\boldsymbol{0.89{\scriptsize \pm 0.02}}$ &
$0.70{\scriptsize \pm 0.03}$ &
$0.63{\scriptsize \pm 0.01}$ &
$0.79{\scriptsize \pm 0.01}$ &
$\boldsymbol{0.66{\scriptsize \pm 0.03}}$ &
$0.63{\scriptsize \pm 0.02}$ &
$0.36{\scriptsize \pm 0.02}$ &
$0.23{\scriptsize \pm 0.01}$ \\
\textbf{C-RADIOv3-ViT-L} & 22 &
$0.54{\scriptsize \pm 0.01}$ &
$0.77{\scriptsize \pm 0.01}$ &
$0.32{\scriptsize \pm 0.02}$ &
$\boldsymbol{0.52{\scriptsize \pm 0.04}}$ &
$\boldsymbol{0.77{\scriptsize \pm 0.02}}$ &
$0.79{\scriptsize \pm 0.01}$ &
$0.74{\scriptsize \pm 0.01}$ &
$0.61{\scriptsize \pm 0.01}$ &
$0.72{\scriptsize \pm 0.02}$ &
$0.60{\scriptsize \pm 0.01}$ &
$0.59{\scriptsize \pm 0.02}$ &
$0.32{\scriptsize \pm 0.03}$ &
$0.19{\scriptsize \pm 0.01}$ \\
\textbf{C-RADIOv3-ViT-L} & 26 &
$0.54{\scriptsize \pm 0.01}$ &
$0.79{\scriptsize \pm 0.01}$ &
$0.31{\scriptsize \pm 0.02}$ &
$0.46{\scriptsize \pm 0.03}$ &
$\underline{0.77{\scriptsize \pm 0.03}}$ &
$0.87{\scriptsize \pm 0.01}$ &
$\underline{0.76{\scriptsize \pm 0.04}}$ &
$0.64{\scriptsize \pm 0.01}$ &
$0.80{\scriptsize \pm 0.02}$ &
$0.61{\scriptsize \pm 0.03}$ &
$\underline{0.60{\scriptsize \pm 0.03}}$ &
$0.32{\scriptsize \pm 0.01}$ &
$\underline{0.25{\scriptsize \pm 0.02}}$ \\
\textbf{C-RADIOv3-ViT-L} & 30 &
$0.54{\scriptsize \pm 0.01}$ &
$0.79{\scriptsize \pm 0.01}$ &
$0.31{\scriptsize \pm 0.02}$ &
$0.46{\scriptsize \pm 0.02}$ &
$0.77{\scriptsize \pm 0.04}$ &
$0.87{\scriptsize \pm 0.01}$ &
$\boldsymbol{0.76{\scriptsize \pm 0.04}}$ &
$0.64{\scriptsize \pm 0.01}$ &
$0.79{\scriptsize \pm 0.01}$ &
$0.61{\scriptsize \pm 0.02}$ &
$\boldsymbol{0.60{\scriptsize \pm 0.03}}$ &
$0.32{\scriptsize \pm 0.01}$ &
$\boldsymbol{0.25{\scriptsize \pm 0.02}}$ \\
\bottomrule
\textbf{C-RADIOv3-ViT-L} & 8-12 &
$0.43{\scriptsize \pm 0.03}$ &
$0.80{\scriptsize \pm 0.03}$ &
$0.29{\scriptsize \pm 0.05}$ &
$0.40{\scriptsize \pm 0.01}$ &
$0.71{\scriptsize \pm 0.02}$ &
$0.83{\scriptsize \pm 0.03}$ &
$0.69{\scriptsize \pm 0.01}$ &
$0.54{\scriptsize \pm 0.01}$ &
$0.77{\scriptsize \pm 0.03}$ &
$0.58{\scriptsize \pm 0.03}$ &
$0.53{\scriptsize \pm 0.02}$ &
$0.32{\scriptsize \pm 0.03}$ &
$0.21{\scriptsize \pm 0.02}$ \\

\textbf{C-RADIOv3-ViT-L} & 12-16 &
$0.49{\scriptsize \pm 0.02}$ &
$0.74{\scriptsize \pm 0.03}$ &
$0.29{\scriptsize \pm 0.03}$ &
$0.43{\scriptsize \pm 0.01}$ &
$0.73{\scriptsize \pm 0.02}$ &
$0.82{\scriptsize \pm 0.01}$ &
$0.73{\scriptsize \pm 0.04}$ &
\underline{$0.65{\scriptsize \pm 0.03}$} &
$0.73{\scriptsize \pm 0.01}$ &
$0.62{\scriptsize \pm 0.02}$ &
$0.52{\scriptsize \pm 0.06}$ &
$0.32{\scriptsize \pm 0.02}$ &
$0.11{\scriptsize \pm 0.02}$ \\

\textbf{C-RADIOv3-ViT-L} & 16-22 &
$0.48{\scriptsize \pm 0.03}$ &
\underline{$0.82{\scriptsize \pm 0.03}$} &
$0.28{\scriptsize \pm 0.03}$ &
$0.46{\scriptsize \pm 0.04}$ &
$0.73{\scriptsize \pm 0.01}$ &
$0.82{\scriptsize \pm 0.05}$ &
$0.73{\scriptsize \pm 0.03}$ &
$0.55{\scriptsize \pm 0.04}$ &
$0.71{\scriptsize \pm 0.02}$ &
$0.57{\scriptsize \pm 0.01}$ &
$0.60{\scriptsize \pm 0.03}$ &
$0.31{\scriptsize \pm 0.03}$ &
$0.19{\scriptsize \pm 0.02}$ \\
\bottomrule
\label{tab:app_layers}
\end{tabular}%
}
\end{table*}

\begin{table*}[t]
\centering
\caption{Comparison on different aligment coefficients, table shows performance across environments (mean $\pm$ SD).}
\setlength{\tabcolsep}{3pt}
\renewcommand{\arraystretch}{1.15}
\resizebox{\textwidth}{!}{%
\begin{tabular}{lcccccccccccccc}
\toprule
\textbf{Teacher} & \textbf{Coeff. (C)} &
\multicolumn{5}{c}{\textbf{Semantic}} &
\multicolumn{5}{c}{\textbf{Vision}} &
\multicolumn{3}{c}{\textbf{Execution}} \\
\cmidrule(lr){3-7}
\cmidrule(lr){8-12}
\cmidrule(lr){13-15}
& & Carrot & Instruct & MultiCarrot & MultiPlate & Plate
& VisionImg & Tex03 & Tex05 & Whole03 & Whole05
& Position & EEPose & PosChangeTo \\
\midrule
\textbf{C-RADIOv3-ViT-L} & 0.2 &
$\boldsymbol{0.61{\scriptsize \pm 0.01}}$ &
$\boldsymbol{0.83{\scriptsize \pm 0.03}}$ &
$\boldsymbol{0.35{\scriptsize \pm 0.02}}$ &
$\boldsymbol{0.49{\scriptsize \pm 0.02}}$ &
$0.75{\scriptsize \pm 0.01}$ &
$\boldsymbol{0.86{\scriptsize \pm 0.02}}$ &
$0.70{\scriptsize \pm 0.02}$ &
$\boldsymbol{0.67{\scriptsize \pm 0.02}}$ &
$\boldsymbol{0.80{\scriptsize \pm 0.02}}$ &
$\boldsymbol{0.60{\scriptsize \pm 0.02}}$ &
$\boldsymbol{0.58{\scriptsize \pm 0.02}}$ &
$\boldsymbol{0.38{\scriptsize \pm 0.02}}$ &
$\boldsymbol{0.20{\scriptsize \pm 0.03}}$ \\
\textbf{C-RADIOv3-ViT-L} & 0.5 &
$0.54{\scriptsize \pm 0.02}$ &
$0.76{\scriptsize \pm 0.01}$ &
$0.32{\scriptsize \pm 0.01}$ &
$0.45{\scriptsize \pm 0.03}$ &
$0.72{\scriptsize \pm 0.02}$ &
$0.80{\scriptsize \pm 0.03}$ &
$0.72{\scriptsize \pm 0.02}$ &
$0.54{\scriptsize \pm 0.02}$ &
$0.74{\scriptsize \pm 0.03}$ &
$0.58{\scriptsize \pm 0.01}$ &
$0.55{\scriptsize \pm 0.03}$ &
$0.35{\scriptsize \pm 0.02}$ &
$0.18{\scriptsize \pm 0.03}$ \\

\textbf{C-RADIOv3-ViT-L} & 1.0 &
$0.51{\scriptsize \pm 0.04}$ &
$0.77{\scriptsize \pm 0.02}$ &
$0.31{\scriptsize \pm 0.05}$ &
$0.47{\scriptsize \pm 0.02}$ &
$\boldsymbol{0.79{\scriptsize \pm 0.03}}$ &
$0.80{\scriptsize \pm 0.02}$ &
$0.70{\scriptsize \pm 0.03}$ &
$0.55{\scriptsize \pm 0.02}$ &
$0.73{\scriptsize \pm 0.01}$ &
$0.61{\scriptsize \pm 0.03}$ &
$0.58{\scriptsize \pm 0.06}$ &
$0.32{\scriptsize \pm 0.07}$ &
$0.14{\scriptsize \pm 0.06}$ \\

\textbf{C-RADIOv3-ViT-L} & 2.0 &
$0.58{\scriptsize \pm 0.03}$ &
$0.78{\scriptsize \pm 0.01}$ &
$0.32{\scriptsize \pm 0.04}$ &
$0.46{\scriptsize \pm 0.08}$ &
$0.79{\scriptsize \pm 0.04}$ &
$\underline{0.86{\scriptsize \pm 0.03}}$ &
$0.69{\scriptsize \pm 0.05}$ &
$0.58{\scriptsize \pm 0.07}$ &
$0.77{\scriptsize \pm 0.04}$ &
$0.61{\scriptsize \pm 0.06}$ &
$0.56{\scriptsize \pm 0.01}$ &
$0.33{\scriptsize \pm 0.03}$ &
$0.16{\scriptsize \pm 0.06}$ \\

\textbf{C-RADIOv3-ViT-L} & 3.0 &
$0.52{\scriptsize \pm 0.10}$ &
$0.78{\scriptsize \pm 0.03}$ &
$0.24{\scriptsize \pm 0.03}$ &
$0.36{\scriptsize \pm 0.02}$ &
$0.73{\scriptsize \pm 0.03}$ &
$0.80{\scriptsize \pm 0.01}$ &
$\boldsymbol{0.79{\scriptsize \pm 0.02}}$ &
$\underline{0.66{\scriptsize \pm 0.05}}$ &
$0.77{\scriptsize \pm 0.04}$ &
$0.57{\scriptsize \pm 0.03}$ &
$0.55{\scriptsize \pm 0.02}$ &
$0.37{\scriptsize \pm 0.03}$ &
$0.15{\scriptsize \pm 0.01}$ \\
\bottomrule
\label{tab:app_coef}
\end{tabular}%
}
\end{table*}

\subsection{Attention maps visualization}
\label{app:att_map}

To further validate the qualitative effect of our alignment objective, we visualize attention maps for Qwen2.5-VL, OpenVLA SFT, and OpenVLA Align (ours) across middle layers of the internal transformer backbone. These layers correspond to the region of strongest vision–language fusion, where attention patterns most directly reflect the model’s visual grounding quality.

\begin{figure}[t]
    \centering
    \includegraphics[width=\linewidth]{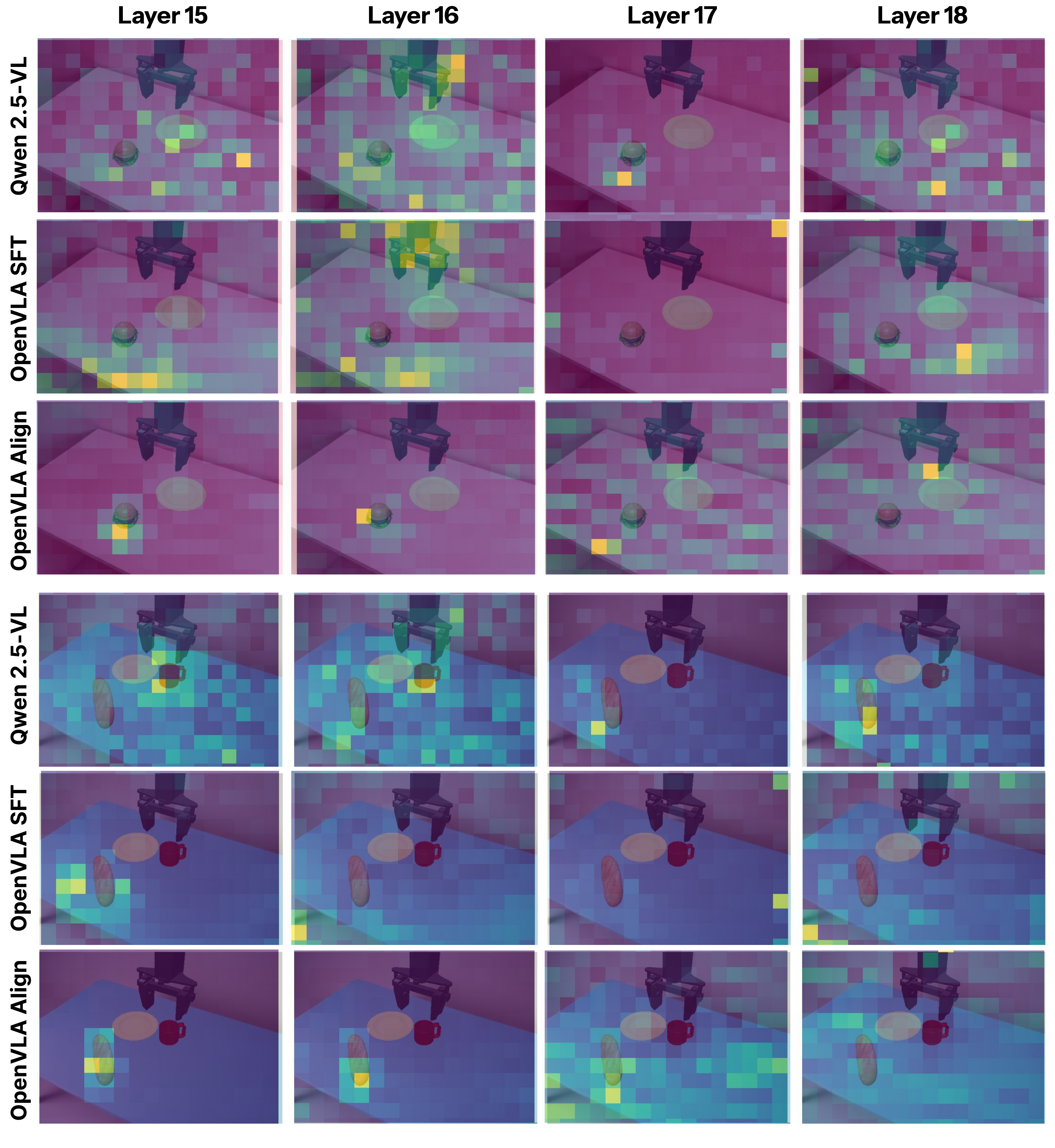}
    \caption{Attention maps across middle layers Qwen2.5-VL, OpenVLA SFT, and OpenVLA Align. 
    The proposed alignment method restores sharp, object-centered attention patterns, improving visual grounding degraded by standard fine-tuning. Question: \textit{"Do you see hamburger|baguette?"}.}
    \label{fig:do_you_see_canv2}
\end{figure}

As shown in \autoref{fig:do_you_see_canv2}, the default OpenVLA SFT exhibits diffuse and spatially inconsistent attention, often extending beyond the queried object. In contrast, our OpenVLA Align model restores sharp, localized focus on task-relevant regions. This confirms that the proposed visual alignment effectively mitigates attention sink introduced by naive fine-tuning and preserves coherent object-centered attention.

\subsection{t-SNE visualization}
\label{app:tsne}

In \autoref{sec:tsne} we show t-SNE \citep{vandermaaten08a} of internal representations for the VLM models and OpenVLA. To keep comparisons strict, we use an out-of-the-box t-SNE implementation with no tuning: perplexity $=30$, max iterations $=1000$, fixed random seed $=42$ (all other parameters at library defaults). These plots are illustrative only (t-SNE distorts global geometry); quantitative conclusions come from linear probing in \autoref{sec:res_lin_prob}.

\subsection{Linear probing}

\begin{table}[H]
\centering
\small
\caption{Linear probing configuration on ImageNet-100 with a frozen backbone and mean-pooled token features.}
\setlength{\tabcolsep}{6pt}
\renewcommand{\arraystretch}{1.15}
\resizebox{0.48\textwidth}{!}{%
\begin{tabular}{lc}
\toprule
\textbf{Parameter} & \textbf{Value} \\
\midrule
Feature pooling & Mean over visual embs. \\ 
Optimizer & SGD (momentum $0.9$) \\
Weight decay & $0.0$ \\
Learning rate & $0.1$ \\
Epochs & $40$ \\
Batch size (train/val/test) & $128$ \\
\bottomrule
\label{tab:linear_probe_imagenet100_config}
\end{tabular}%
}
\end{table}

For reproducibility and fair comparison, we evaluate representational quality with a frozen-feature linear probe under a single, fixed configuration (see \autoref{tab:linear_probe_imagenet100_config}). Concretely, we extract patch embeddings (mean-pooled to a single vector) from the final C-RADIOv3 teacher and from intermediate visual layers of each OpenVLA variant. A single linear classifier is trained on top of these frozen features. All hyperparameters are held constant across models and layers, and the same random seed and data split are used for every run. Due to computational constraints, we operate on a reduced ImageNet-100. We report top-1 accuracy on the evaluation split, without per-any tuning, so any differences reflect only the underlying representations rather than linear probe tuning changes.

\subsection{Ablations}
\label{app:ablations}

This section provides the complete ablation results that underlie the analyses in \autoref{sec:ablations}, Vision teachers (\autoref{tab:app_abb_vision_teacher}),
Alignment projectors (\autoref{tab:app_abb_vision_projector}),
Alignment layers (\autoref{tab:openvla_train_config_alig}), Alignment coefficients (\autoref{tab:app_coef}). In the ablation studies presented in \autoref{sec:ablations}, we test the hypothesis that a given model variant (denoted B) outperforms the baseline variant (A) in terms of success rate. For each pairwise comparison, we fix all other parameters altering only the component under investigation (e.g., alignment objective, layer depth, projection type). This ensures that any observed performance difference can be attributed solely to the ablated design choice.

To assess statistical significance, we use the paired Wilcoxon signed-rank test \citep{wilcoxon1945individual}, a non-parametric test suited for comparing two matched samples that do not necessarily follow a normal distribution. The unit of analysis is the per-seed success rate over matched trials, evaluated independently for each environment type (Semantic, Vision, Execution). The test uses a one-sided alternative hypothesis ($H_1$: B > A), corresponding to our directional research question and report the exact p-value. All comparisons are conducted over 128 shared random seeds, ensuring that each seed–environment pair is identical across the methods being compared. This careful experimental control allows us to draw meaningful conclusions about the contribution of each individual design choice.

\subsection{Different projection approaches}
Below, we provide detailed formulations of the various projectors used in \autoref{sec:proj_type} of our experiments.

\paragraph{\textbf{Cosine Projector.}}
A normalized linear projection that preserves angular similarity:
\begin{equation}
z = \frac{W h}{\|W h\|_2}, \qquad 
W \in \mathbb{R}^{d_z \times d_{\text{hidden}}}.
\end{equation}

\paragraph{\textbf{Orthogonal Projector.}}
A fixed linear transform with orthonormal columns:
\begin{equation}
W^\top W = I_{d_z}, \quad z = W h.
\end{equation}

\paragraph{\textbf{Random Fourier Feature (RFF) Projector.}}
A fixed randomized mapping that implicitly approximates a kernel feature space:
\begin{equation}
z = \sqrt{\frac{2}{D}} 
\cos(W h + b), \quad 
W_{ij} \sim \mathcal{N}(0, \gamma^{-2}),\;
b_i \sim \mathcal{U}(0, 2\pi).
\end{equation}

\paragraph{\textbf{Whitening–Affine Projector.}}
Combines feature whitening and affine normalization:
\begin{equation}
z = \Lambda^{-1/2}(h - \mu) + b,
\end{equation}

\paragraph{\textbf{Spectral–Norm Projector.}}
A constrained linear mapping enforcing bounded operator norm:
\begin{equation}
z = W h, \qquad \|W\|_2 \leq 1.
\end{equation}

\end{document}